\documentclass[10pt,twocolumn]{article}

\usepackage{wacv}
\usepackage{times}
\usepackage{epsfig}
\usepackage{graphicx}
\usepackage{amsmath}
\usepackage{amssymb}
\usepackage{array}
\usepackage{bbm}
\usepackage{amsthm}
\usepackage{xspace}
\usepackage{enumitem}

\usepackage[pagebackref=true,breaklinks=true,colorlinks,bookmarks=false]{hyperref}

\wacvfinalcopy

\ifwacvfinal\fi

\usepackage{tikzextern}
\newcommand{\extfig}[2]{\tikzsetnextfilename{figs/extern/#1}{#2}}
\newcommand{\extdata}[1]{}

\begin{document}

\title{Unsupervised object discovery for instance recognition}

\author{
Oriane Sim\'eoni$^1$ \ \ \ \ Ahmet Iscen$^2$ \ \ \ \ Giorgos Tolias$^2$\ \ \ \ Yannis Avrithis$^{1}$\ \ \ \ Ond{\v r}ej Chum$^{2}$\\
{\fontsize{11}{13}\selectfont$^1$Inria Rennes\ \ \ \ \ \ $^2$VRG, FEE, CTU in Prague}\\
{\fontsize{10}{11}\selectfont \texttt{\{oriane.simeoni,ioannis.avrithis\}@inria.fr}} \\
{\fontsize{10}{11}\selectfont \texttt{\{ahmet.iscen,giorgos.tolias,chum\}@cmp.felk.cvut.cz}}
}

\maketitle

\begin{abstract}
Severe background clutter is challenging in many computer vision tasks, including large-scale image retrieval. Global descriptors, that are popular due to their memory and search efficiency, are especially prone to corruption by such a clutter. Eliminating the impact of the clutter on the image descriptor increases the chance of retrieving relevant images and prevents topic drift due to actually retrieving the clutter in the case of query expansion. In this work, we propose a novel salient region detection method. It captures, in an unsupervised manner, patterns that are both discriminative and common in the dataset. Saliency is based on a centrality measure of a nearest neighbor graph constructed from regional CNN representations of dataset images. The descriptors derived from the salient regions improve particular object retrieval, most noticeably in a large collections containing small objects.
\end{abstract}

\newcommand{\FSexh}{FS.\textsc{exh}\xspace}
\newcommand{\FSegm}{FS.\textsc{egm}\xspace}
\newcommand{\OSexh}{OS.\textsc{exh}\xspace}
\newcommand{\OSegm}{OS.\textsc{egm}\xspace}
\newcommand{\cutoff}{\tau}
\newcommand{\salPow}{\rho}
\newcommand{\scale}{\sigma}
\newcommand{\dbPow}{\theta}
\newcommand{\queryPow}{\Theta}
\newcommand{\FS}{FS\xspace}
\newcommand{\OS}{OS\xspace}

\setlist[enumerate]{topsep=4pt,parsep=0pt,partopsep=0pt,itemsep=2pt}
\setlist[itemize]{topsep=4pt,parsep=0pt,partopsep=0pt,itemsep=2pt}

\newcommand{\head}[1]{{\noindent\bf #1}}
\newcommand{\red}[1]{{\color{red}{#1}}}
\newcommand{\blue}[1]{{\color{blue}{#1}}}
\newcommand{\green}[1]{{\color{green}{#1}}}
\newcommand{\gray}[1]{{\color{gray}{#1}}}

\newcommand{\ind}{\mathbbm{1}}
\newcommand{\expect}{\mathbb{E}}
\newcommand{\nat}{\mathbb{N}}
\newcommand{\zahl}{\mathbb{Z}}
\newcommand{\real}{\mathbb{R}}
\def\l2{\ensuremath{\ell_2}\xspace}

\newcommand{\T}{{\!\top}}
\newcommand{\mT}{{-\!\top}}

\newcommand{\diag}{\operatorname{diag}}
\newcommand{\defn}{\mathrel{\operatorname{:=}}}
\newcommand{\norm}[1]{\left\|{#1}\right\|}
\newcommand{\ceil}[1]{\left\lceil{#1}\right\rceil}
\newcommand{\inner}[2]{\left\langle{#1},{#2}\right\rangle}

\newcommand{\wb}[1]{\overline{#1}}
\newcommand{\wt}[1]{\widetilde{#1}}

\def\xssp{\hspace{1pt}}
\def\ssp{\hspace{3pt}}
\def\msp{\hspace{5pt}}
\def\lsp{\hspace{12pt}}

\newcommand{\cA}{\mathcal{A}}
\newcommand{\cB}{\mathcal{B}}
\newcommand{\cC}{\mathcal{C}}
\newcommand{\cD}{\mathcal{D}}
\newcommand{\cE}{\mathcal{E}}
\newcommand{\cF}{\mathcal{F}}
\newcommand{\cG}{\mathcal{G}}
\newcommand{\cH}{\mathcal{H}}
\newcommand{\cI}{\mathcal{I}}
\newcommand{\cJ}{\mathcal{J}}
\newcommand{\cK}{\mathcal{K}}
\newcommand{\cL}{\mathcal{L}}
\newcommand{\cM}{\mathcal{M}}
\newcommand{\cN}{\mathcal{N}}
\newcommand{\cO}{\mathcal{O}}
\newcommand{\cP}{\mathcal{P}}
\newcommand{\cQ}{\mathcal{Q}}
\newcommand{\cR}{\mathcal{R}}
\newcommand{\cS}{\mathcal{S}}
\newcommand{\cT}{\mathcal{T}}
\newcommand{\cU}{\mathcal{U}}
\newcommand{\cV}{\mathcal{V}}
\newcommand{\cW}{\mathcal{W}}
\newcommand{\cX}{\mathcal{X}}
\newcommand{\cY}{\mathcal{Y}}
\newcommand{\cZ}{\mathcal{Z}}

\newcommand{\vA}{\mathbf{A}}
\newcommand{\vB}{\mathbf{B}}
\newcommand{\vC}{\mathbf{C}}
\newcommand{\vD}{\mathbf{D}}
\newcommand{\vE}{\mathbf{E}}
\newcommand{\vF}{\mathbf{F}}
\newcommand{\vG}{\mathbf{G}}
\newcommand{\vH}{\mathbf{H}}
\newcommand{\vI}{\mathbf{I}}
\newcommand{\vJ}{\mathbf{J}}
\newcommand{\vK}{\mathbf{K}}
\newcommand{\vL}{\mathbf{L}}
\newcommand{\vM}{\mathbf{M}}
\newcommand{\vN}{\mathbf{N}}
\newcommand{\vO}{\mathbf{O}}
\newcommand{\vP}{\mathbf{P}}
\newcommand{\vQ}{\mathbf{Q}}
\newcommand{\vR}{\mathbf{R}}
\newcommand{\vS}{\mathbf{S}}
\newcommand{\vT}{\mathbf{T}}
\newcommand{\vU}{\mathbf{U}}
\newcommand{\vV}{\mathbf{V}}
\newcommand{\vW}{\mathbf{W}}
\newcommand{\vX}{\mathbf{X}}
\newcommand{\vY}{\mathbf{Y}}
\newcommand{\vZ}{\mathbf{Z}}

\newcommand{\va}{\mathbf{a}}
\newcommand{\vb}{\mathbf{b}}
\newcommand{\vc}{\mathbf{c}}
\newcommand{\vd}{\mathbf{d}}
\newcommand{\ve}{\mathbf{e}}
\newcommand{\vf}{\mathbf{f}}
\newcommand{\vg}{\mathbf{g}}
\newcommand{\vh}{\mathbf{h}}
\newcommand{\vi}{\mathbf{i}}
\newcommand{\vj}{\mathbf{j}}
\newcommand{\vk}{\mathbf{k}}
\newcommand{\vl}{\mathbf{l}}
\newcommand{\vm}{\mathbf{m}}
\newcommand{\vn}{\mathbf{n}}
\newcommand{\vo}{\mathbf{o}}
\newcommand{\vp}{\mathbf{p}}
\newcommand{\vq}{\mathbf{q}}
\newcommand{\vr}{\mathbf{r}}
\newcommand{\Vs}{\mathbf{s}}
\newcommand{\vt}{\mathbf{t}}
\newcommand{\vu}{\mathbf{u}}
\newcommand{\vv}{\mathbf{v}}
\newcommand{\vw}{\mathbf{w}}
\newcommand{\vx}{\mathbf{x}}
\newcommand{\vy}{\mathbf{y}}
\newcommand{\vz}{\mathbf{z}}

\def\sssp{\hspace{1pt}}
\def\ssp{\hspace{3pt}}
\def\msp{\hspace{5pt}}
\def\bsp{\hspace{12pt}}

\newcommand{\os}[1]{\textbf{#1}}
\newcommand{\ns}[1]{{\textbf{\textcolor{red}{#1}}}}
\newcommand{\alert}[1]{{\color{red}{#1}}}

\def \sim{s}
\def \x{\mathbf{x}}
\def \z{\mathbf{z}}

\newcommand{\vone}{\mathbf{1}}
\newcommand{\vzero}{\mathbf{0}}

\newcommand{\vmu}{\boldsymbol{\mu}}

\newcommand{\rLambda}{\mathrm{\Lambda}}
\newcommand{\rSigma}{\mathrm{\Sigma}}

\newcommand{\mypar}[1]{\noindent \textbf{#1}}

\makeatletter
\newcommand*\bdot{\mathpalette\bdot@{.7}}
\newcommand*\bdot@[2]{\mathbin{\vcenter{\hbox{\scalebox{#2}{$\m@th#1\bullet$}}}}}
\makeatother

\newenvironment{narrow}[1][1pt]
	{\setlength{\tabcolsep}{#1}}
	{\setlength{\tabcolsep}{6pt}}

\makeatletter
\DeclareRobustCommand\onedot{\futurelet\@let@token\@onedot}
\def\@onedot{\ifx\@let@token.\else.\null\fi\xspace}
\def\eg{\emph{e.g}\onedot} \def\Eg{\emph{E.g}\onedot}
\def\ie{\emph{i.e}\onedot} \def\Ie{\emph{I.e}\onedot}
\def\cf{\emph{c.f}\onedot} \def\Cf{\emph{C.f}\onedot}
\def\etc{\emph{etc}\onedot} 
\def\wrt{w.r.t\onedot} \def\dof{d.o.f\onedot}
\def\etal{\emph{et al}\onedot}
\makeatother

\section{Introduction}
\label{sec:intro}

Particular object retrieval becomes very challenging when the object of interest is covering a small part of the image. In this case, the amount of relevant information is significantly reduced. Large objects might be partially occluded, while small objects are on a background that covers most of the image. A combination of both, occlusion and cluttered background, is not rare either. These conditions naturally arise from image acquisition and make naive approaches fail, including global template matching or
semi-robust template matching~\cite{OT06}.

\begin{figure}
\vspace{-5pt}
\setlength{\fboxsep}{0pt}%
\setlength{\fboxrule}{1.5pt}%
\newcommand{\showIm}[3]{\includegraphics[width=1.6cm,height=1.6cm]{figs/examples/#1_q_#2_#3.jpg}}
\newcommand{\qIm }[2]{\includegraphics[width=1.6cm,height=1.6cm]{figs/examples/{#2}_q_#1_bbx.jpg}}
\newcommand{\pr}[2]{\footnotesize{{\color{red}#1}{\scriptsize$\rightarrow$}{\color{blue}#2}}}

\newcommand{\posEx}[1]{\footnotesize{\color{blue}#1}}
\newcommand{\negEx}[1]{\footnotesize{\color{red}#1}}

\begin{center} \footnotesize
   \begin{tabular}
   {*{3}{@{\msp}c@{\msp}}}

\includegraphics[height=4.0cm]{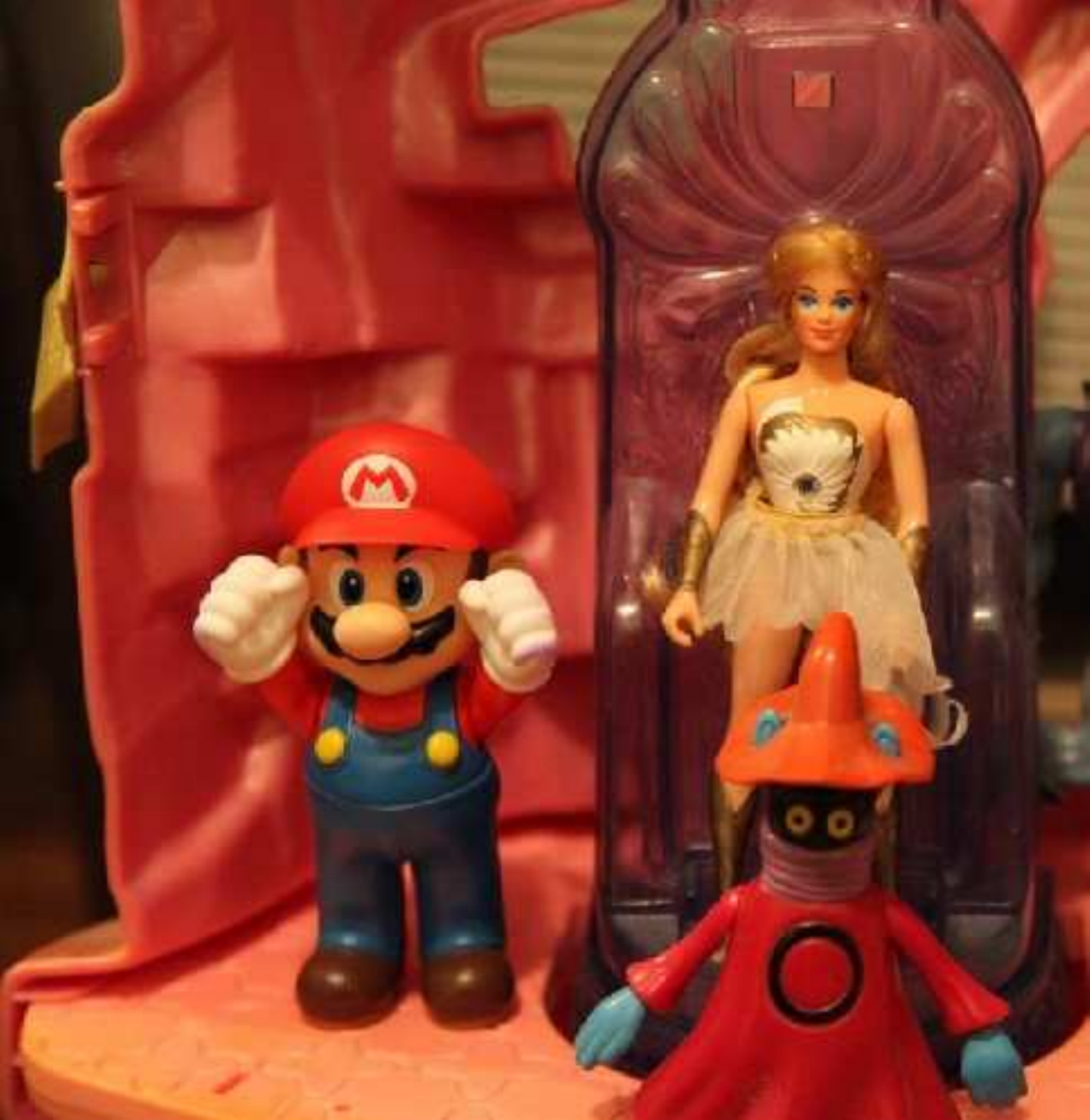} &
\includegraphics[height=4.0cm]{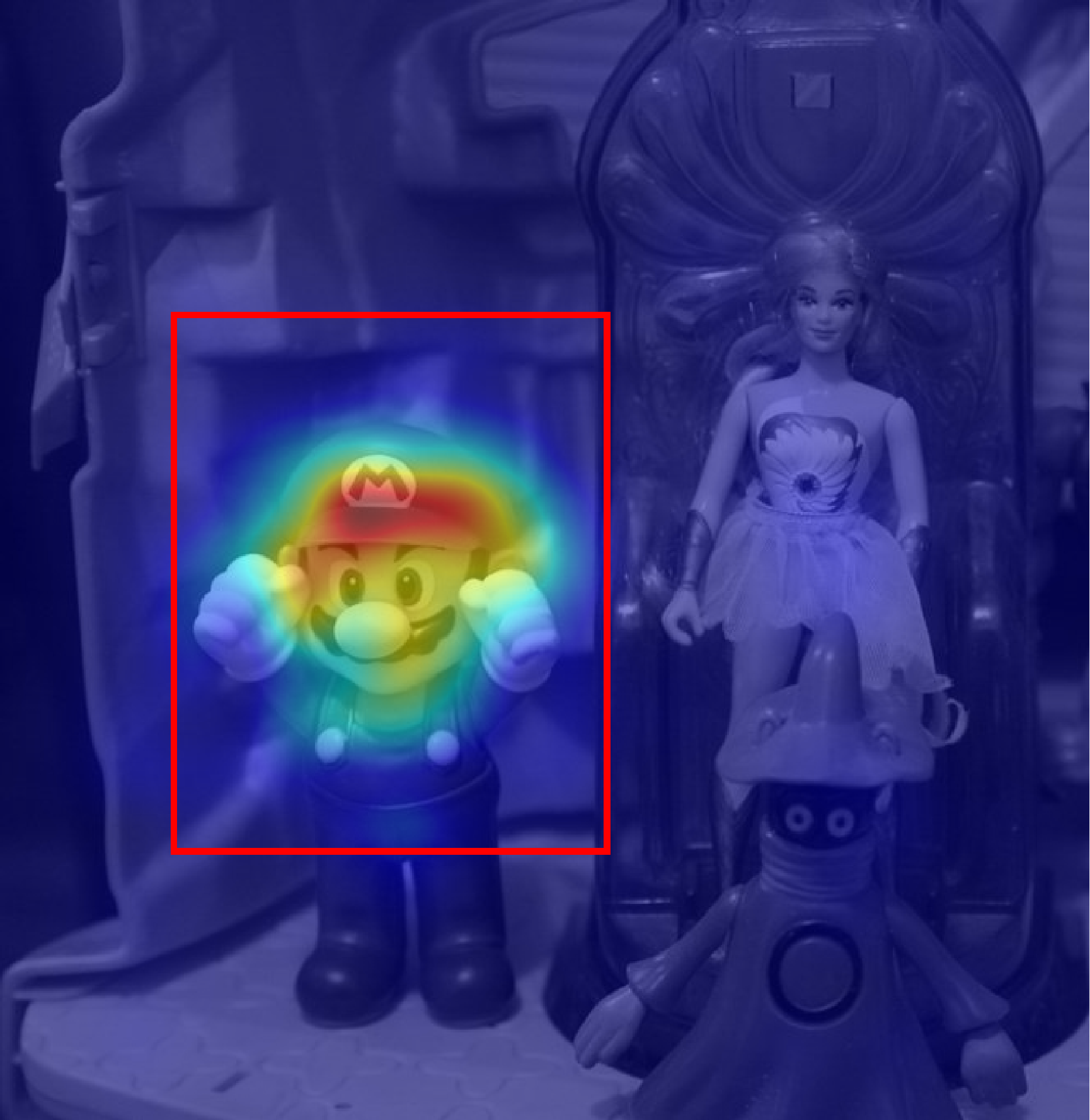} \\
&&\\[-8pt]
\\[-2pt]

  \end{tabular}
\end{center}
 \vspace{-20pt}

\caption{The saliency map (right) computed for an input image (left) based on common-structure analysis on \emph{Instre} dataset. Background clutter and objects not relevant for this dataset are automatically removed. The image is represented only by the region detected on the saliency map.}

\label{fig:intro}
 \vspace{-10pt}
\end{figure}

Ideally, image descriptors should be extracted only from the relevant part of the image, suppressing the irrelevant clutter and occlusions. In this paper, we attempt to determine the regions containing the relevant information, as shown in Figure~\ref{fig:intro}, in a fully unsupervised manner.

Methods based on robust matching of \emph{hand-crafted local features} are naturally insensitive to occlusion and background clutter. The locality of the features allows to match small parts of images in regions containing the object of interest, while the incorrect matches are typically removed by robust geometric consistency check~\cite{PCISZ07}. Methods based on efficient matching of vector-quantized local-feature descriptors were introduced in context of image retrieval by Sivic and Zisserman~\cite{SZ03}.

Retrieval methods based on descriptors extracted by \emph{convolutional neural networks} (CNNs) have become popular because they combine good precision and recall, efficiency of the search, and reasonable memory footprint~\cite{BSCL14,RSAC14}. Deep neural networks are capable of learning, to some extent, what information in the image is relevant, which results in a good performance even with global descriptors~\cite{TSJ15,BL15,KMO15}. However, if the signal to noise ratio is low, \eg the object is relatively small, multiple objects are present, \etc, the global CNN descriptors fail~\cite{ITA+16,IAT+17}.

A class of methods inspired by \emph{object detection} have recently emerged. Instead of attempting to match the whole image to the query, the problem is changed to finding a rectangular region in the image that best matches the query~\cite{TSJ15,SGMS16}. An inefficient search by sliding window is intractable for large collections of images.  The exhaustive enumeration is approximated by similarity evaluation on a number of pre-selected regions. The regions are either selected geometrically to cover the whole image at different scales, as in R-MAC~\cite{TSJ15}, or by considering the content by object or region proposal methods~\cite{SGMS16,SHGXS17,GARL16}.

Another direction of suppressing irrelevant content is saliency detection~\cite{KMO15,NASH16}. For each image, a saliency map, that captures more general region shapes compared to (a small set of) rectangles, is first estimated. The contribution of each pixel (or region) is then proportional to the saliency of that location.

In this work we introduce a very simple pooling scheme that inherits the properties of both saliency detection and region based pooling and that, like all previous approaches, is applied to each image in the database \emph{independently}. In addition, we investigate the use of the resulting regional representation for automatic, offline object discovery and suppression of background clutter, which considers the image collection \emph{as a whole}. Unlike previous approaches, we do this in an unsupervised way.
As a consequence, our representation takes two saliency detection steps into account. One that acts per image and depends solely on its content and another that considers the image collection as a whole and captures frequently appearing objects.

In both cases, we derive a \emph{global} representation that outperforms comparable state-of-the-art methods in retrieving small objects on standard benchmarks, while the memory footprint and online cost is only a fraction compared to more powerful \emph{regional} representations~\cite{RSAC14,ITA+16}. Moreover, we show that our representation benefits significantly from \emph{query expansion} methods.

Section~\ref{sec:related} discusses our contributions against related work. Section~\ref{sec:method} describes our methodology including our pooling scheme in Section~\ref{sec:crow} and our object discovery approach in Section~\ref{sec:saliency}. We present experimental results in Section~\ref{sec:exp} and draw conclusions in Section~\ref{sec:discussion}.

\section{Related work}
\label{sec:related}
Local features and geometric matching offer an attractive way for retrieval systems to handle occlusions, clutter, and small objects~\cite{SZ03,PCISZ07,JDS10a}.
One of their drawbacks is high query complexity and large storage cost; an image is typically represented by several thousands features.
Many methods attempt to decrease the amount of indexed features by removing background clutter while maintaining the relevant information.
The selection procedure is either applied independently per image or considers an image collection as a whole.
Common examples of the former case are bursty feature detection~\cite{SAJ15}, symmetry detection~\cite{TKA12} or use of semantic segmentation~\cite{AZ14b,OPTA+08}.
The methods of the second category, are scalable enough to jointly process the whole collection and perform feature selection by the following assumption. A feature that repeats over multiple instances of the same object in the dataset is likely to appear in novel views of the object too.
Representative cases are  common object discovery~\cite{TL09,TAJ15}, co-occurrence detection~\cite{CM10}, or methods using GPS information~\cite{GBQG09,KSP10}.

The work by Turcot and Lowe~\cite{TL09} performs pairwise spatial verification on hand-crafted local features across all images and only indexes verified features.
With an additional off-line cost, the on-line stage is sped up and the memory footprint is reduced.
However, unique views of objects are not verified and thus discarded.
In this work, we address a similar selection problem based on more powerful CNN-based representation rather than local features. 

Recent advances on deep learning~\cite{ARSM+14,TSJ15,KMO15,GARL16b,RTC16} dispense with the large memory footprint by using global descriptors and cast the problem of instance search as Euclidean nearest neighbor search.
Nevertheless, background clutter and occlusion are better handled by regional representation.
Regional descriptors significantly increase the performance when they are indexed independently~\cite{RSAC14,ITA+16} but this comes at a prohibited memory and computational cost for large scale scenarios. 
Region Proposal Networks (RPN) are applied either off-the-shelf~\cite{SGMS16} or after fine-tuning~\cite{SHGXS17} for instance search.
The RPNs reduce the number of regions per image only to the order of tens.
Our work focuses on aggregating regional representation that keeps the complexity low but we rather detect regions around salient objects and objects that frequently appear in the dataset.  
Jimenez~\etal~\cite{JAG17} construct saliency maps and perform region detection to construct global image vectors, as we also do.
However, they employ generic object detectors trained on ImageNet and this makes the method not applicable with fine-tuned networks which provide the best performance.  
The Hessian-affine detector is used on CNN activations to detect repeatable regions~\cite{JCSC17}. 
The major benefit in this work, though, comes from second order pooling and higher dimensional descriptors.

Saliency maps are another way to handle clutter and occlusions. 
Once more, there exist both examples of computation in an unsupervised manner~\cite{KMO15,LK17} or learned~\cite{NASH16,JDF17} and applied per image afterwards.
Our approach generates saliency maps in a fully unsupervised way that capture both salient objects on single images but also repeating objects appearing in a particular image  collection.

\section{Method}
\label{sec:method}
Like~\cite{TL09}, our objective is to remove transient and non-distinctive objects as in Figure~\ref{fig:intro} and rather focus on objects appearing frequently in a dataset. Beginning with the activation map of a convolutional layer in a CNN, one would need access to a local representation to automatically discover such objects. On the other hand, knowing what these objects are would help forming a local representation by selecting regions depicting them, which appears to be a chicken-and-egg problem. Without an initial region selection, we risk ``discovering'' uninformative but frequently appearing ``stuff''-like patches, for instance sky.

\subsection{Overview}
\label{sec:overview}
\input{tex/fig_overview}
Fortunately, it is possible to make an initial selection based on CNN activations alone, without any training and without bounding box annotations. As described in Section~\ref{sec:crow}, the mechanism is inspired by CroW~\cite{KMO15} and Grad-CAM~\cite{SDV+16} and generates a \emph{feature saliency} map. This initiates our offline analysis illustrated in Figure~\ref{fig:overview}. A small set of rectangular regions is detected per image from this map as discussed in Section~\ref{sec:egm}. This first round of detection is applied independently per image and depends only on its content. 

Each region in the dataset is associated to a feature saliency score and a visual descriptor, pooled from the activation map of the corresponding image, as discussed in Section~\ref{sec:desc}. It is now possible to compute a \emph{centrality} score per region, representing the ``significance'' of each region in the dataset. This is based on a region $k$-NN graph and is discussed in Sections~\ref{sec:graph} and \ref{sec:katz}.

Now, given a new image, we can infer the ``significance'' of every region from its nearest neighbors in the graph, yielding a dense \emph{object saliency} map as discussed in Section~\ref{sec:saliency}. This is a regression problem and we suggest a non-parametric $k$-NN solution. Finally, we detect a small set of rectangular regions on this saliency map and extract a global descriptor to represent dataset images for retrieval, as discussed in Section~\ref{sec:repr}. This second detection procedure takes into account all salient and repeating objects appearing in the dataset.

The entire process is fully unsupervised and only assumes on the-shelf networks trained on a classification or retrieval task without bounding box annotations.

\subsection{Representation}
\label{sec:notation}
We represent the activation map of a convolutional layer as a non-negative 3d tensor $A \in \real^{h \times w \times c}$ where $h, w$ are the spatial resolution (height, width) and $c$ is the number of feature channels. The set of valid spatial positions is $P \defn [h] \times [w]$\footnote{Here, $[i]$ is the set $\{1, \dots, i\}$ for $i \in \nat$.} and the set of all rectangles with vertices in $P$ is denoted by $\cR$. By $A_{pj}$ we represent an element of $A$ at position $p \in P$ and channel $j \in [c]$. By $A_{\bdot j} \in \real^{h \times w}$ we denote the 2d feature map of $A$ corresponding to channel $j \in [c]$. By $A_{p \bdot} \in \real^c$ we denote the vector containing all feature channels at position $p \in P$.

\subsection{Feature saliency}
\label{sec:crow}

Inspired by \emph{cross-dimensional weighting and pooling} (CroW)~\cite{KMO15} and \emph{class activation mapping} (CAM)~\cite{ZKL+16}, we construct a 2d saliency map of an image based on a convolution activation of that image alone. Following CroW, we compute an idf-like weight per channel $\vb \in \real^c$ with elements
\begin{equation}
	b_j = \log \left( \frac{(\va + \epsilon)^\T \vone}{a_j + \epsilon} \right)
	\label{eq:idf}
\end{equation}
for $j \in [c]$, where $\va \defn \frac{1}{wh} \sum_{p \in P} \ind [A_{p \bdot}] \in \real^c$ is the average number of nonzero elements per channel. We then compute a weighted sum over channels
\begin{equation}
	F \defn \sum_{j \in [c]} b_j A_{\bdot j}
	\label{eq:feat}
\end{equation}
Finally, we obtain the 2d \emph{feature saliency} (\FS) map $\hat{F} \in \real^{h \times w}$ by normalizing $F$ according to~\cite{KMO15}. 
Contrary to CroW, we use the feature channel weights when computing the 2d spatial weights, amplifying channels with sparse activation. This order of summation is the same as in CAM. However, we are working with channel weights obtained by a sparsity property on any convolutional layer, without any assumption on the network topology. CAM on the other hand, assumes global average pooling followed by a fully connected layer mapping channels to classes and uses the parameters of this layer to obtain a saliency map per class.

\subsection{Region detection}
\label{sec:egm}
\newcommand{\egmfs}[1]{\includegraphics[height=3cm]{figs/egm/fs/#1}}

\begin{figure*}
\begin{center}
\begin{narrow}
\begin{tabular}{ccccccc}
\egmfs{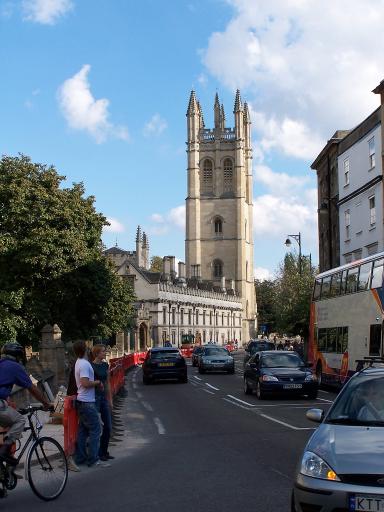} &
\egmfs{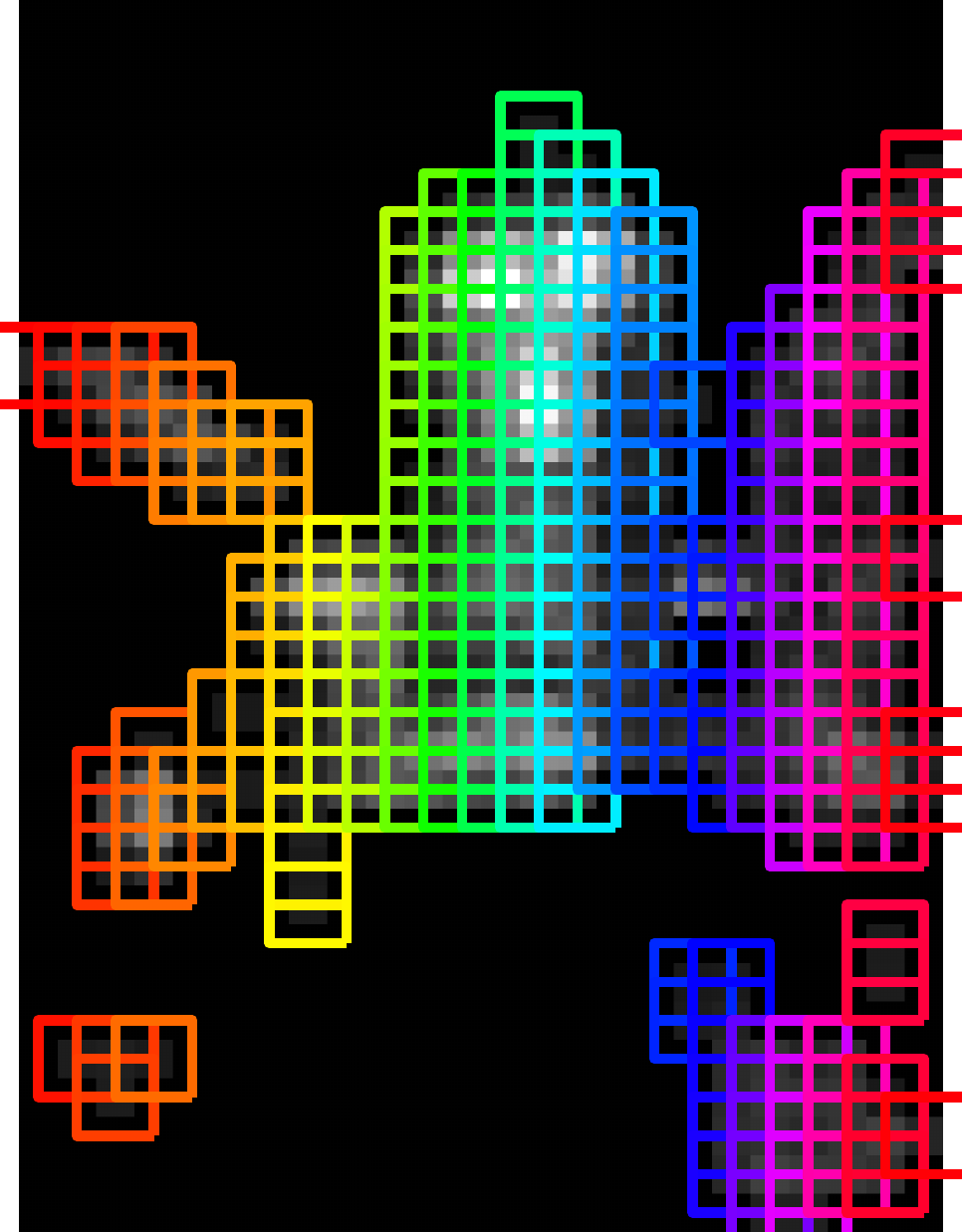} & \egmfs{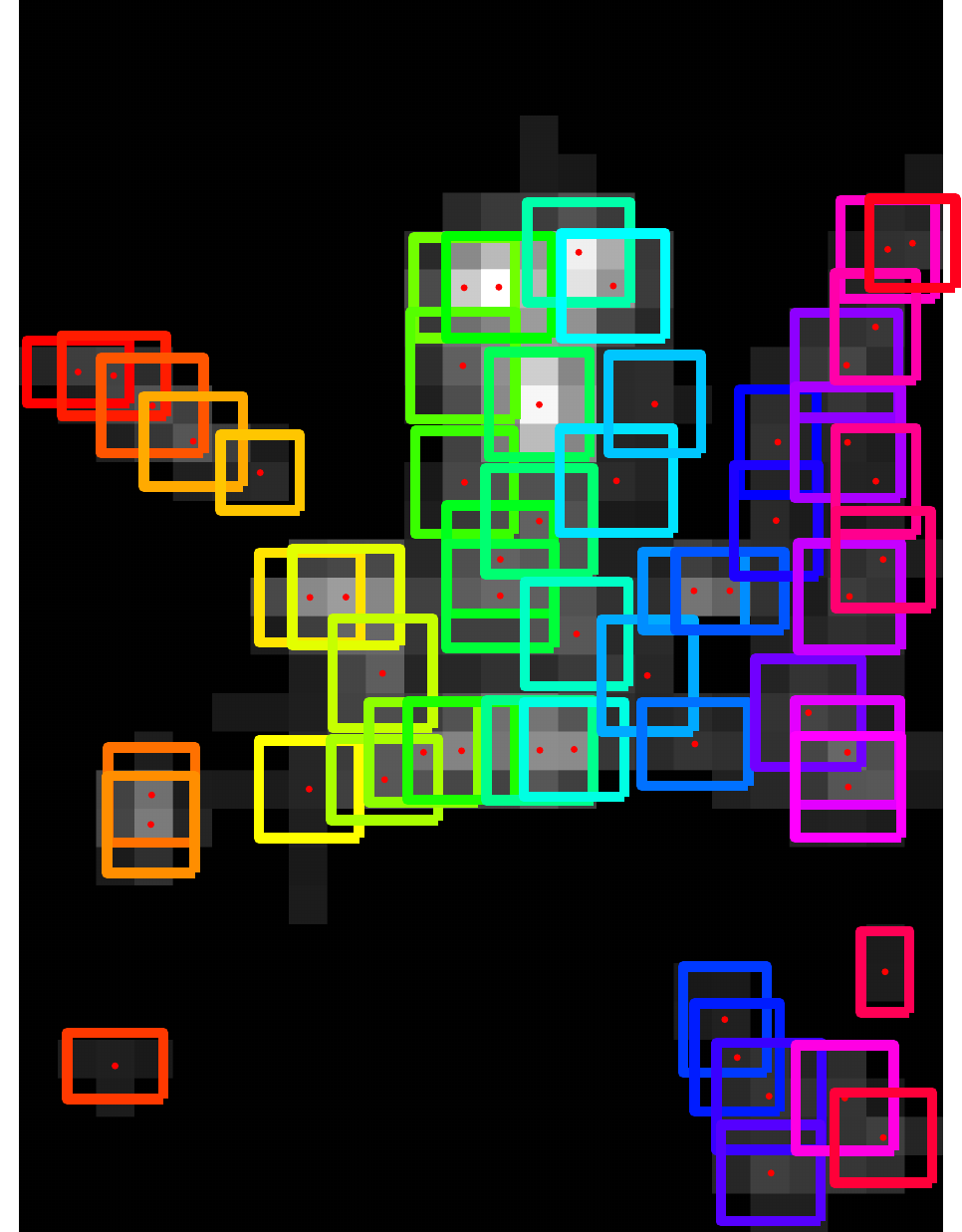} & \egmfs{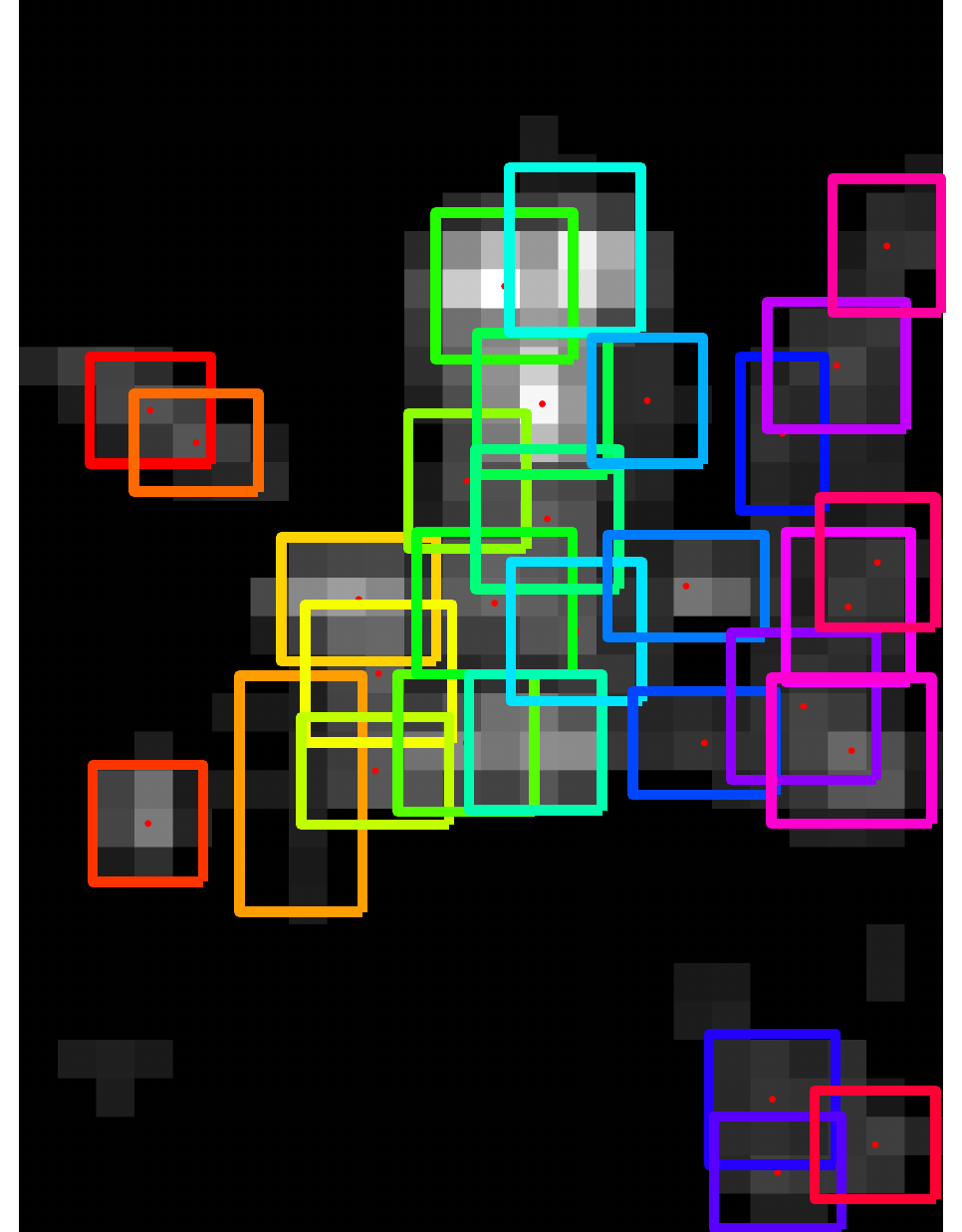} & \egmfs{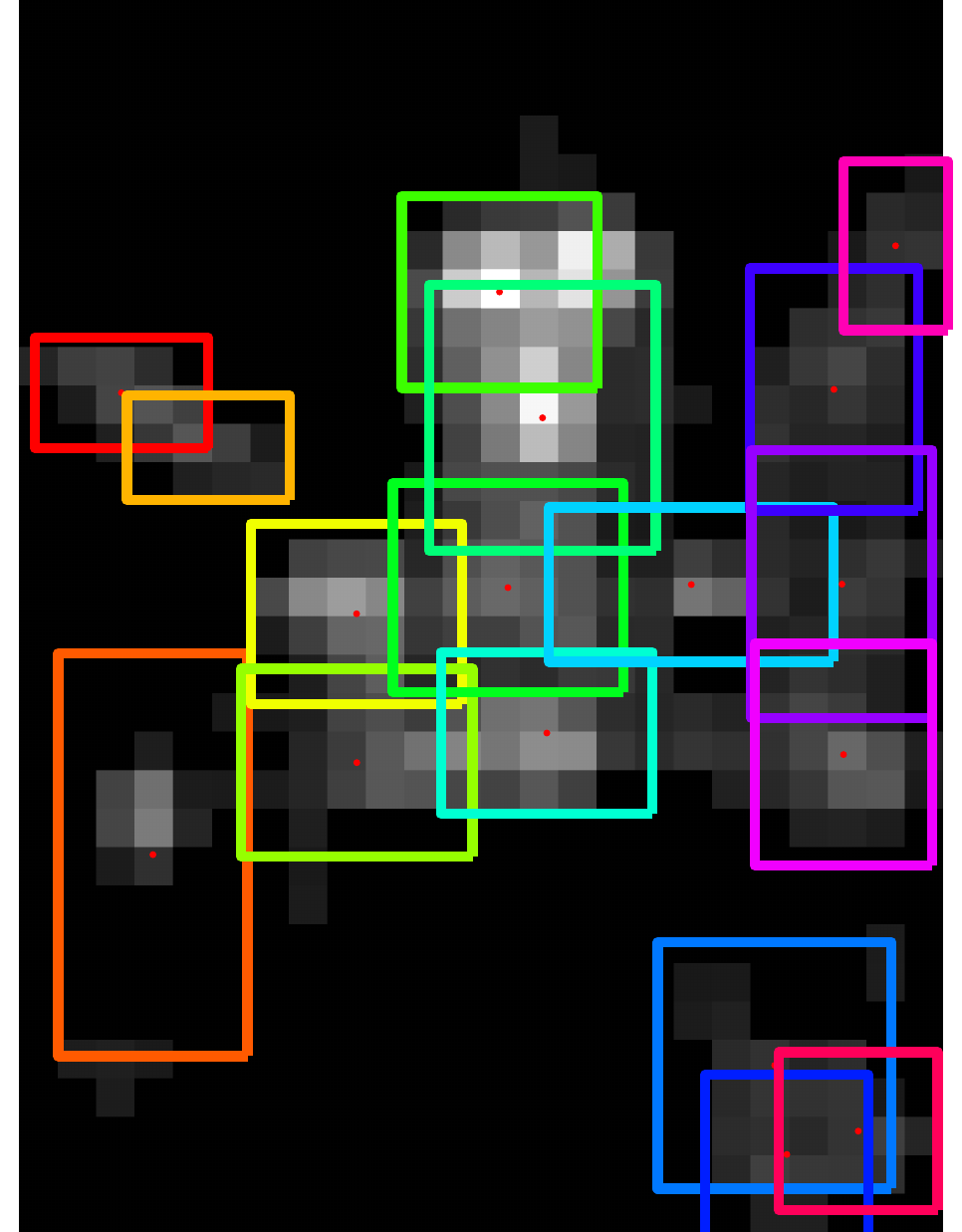} &  \egmfs{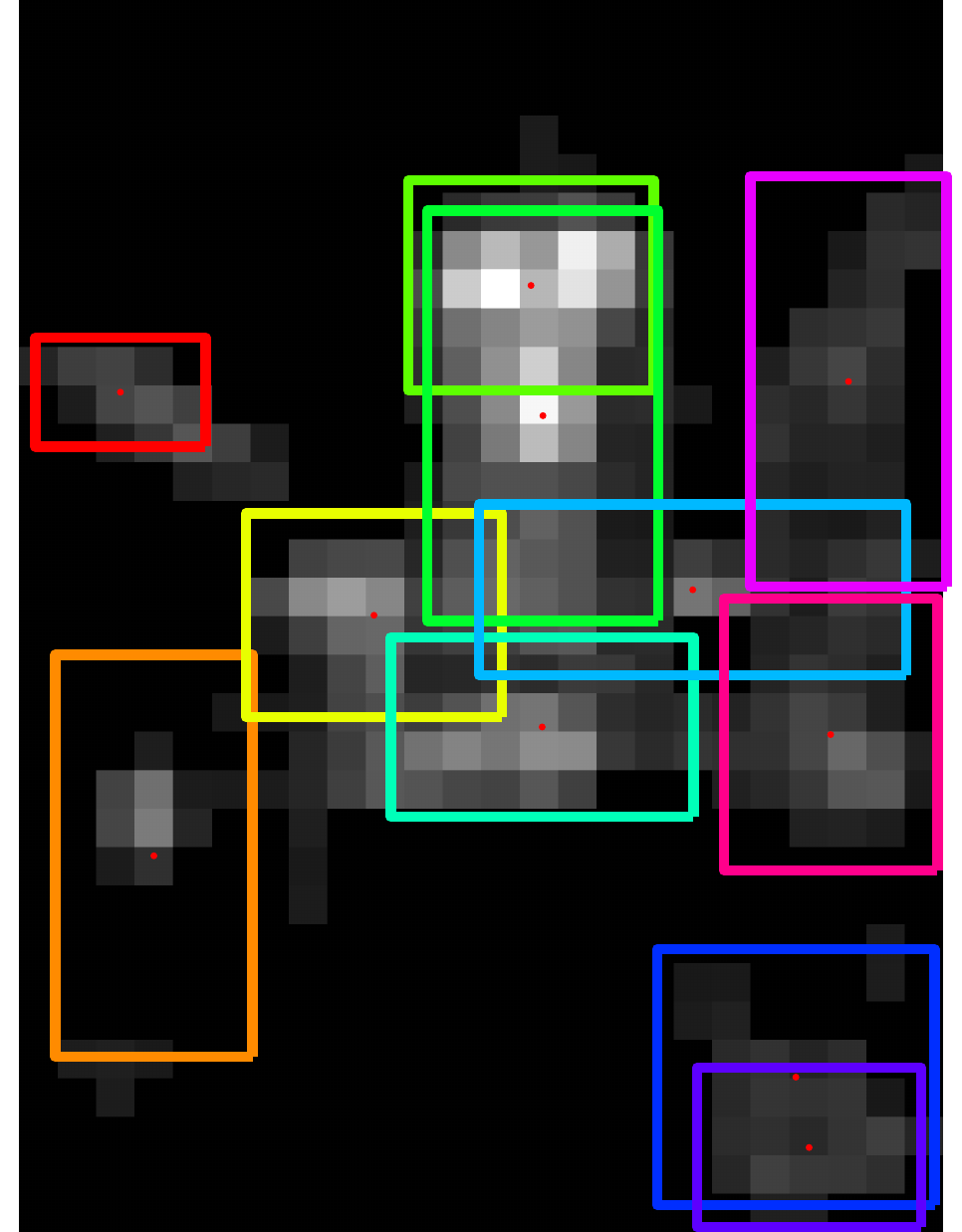} & \egmfs{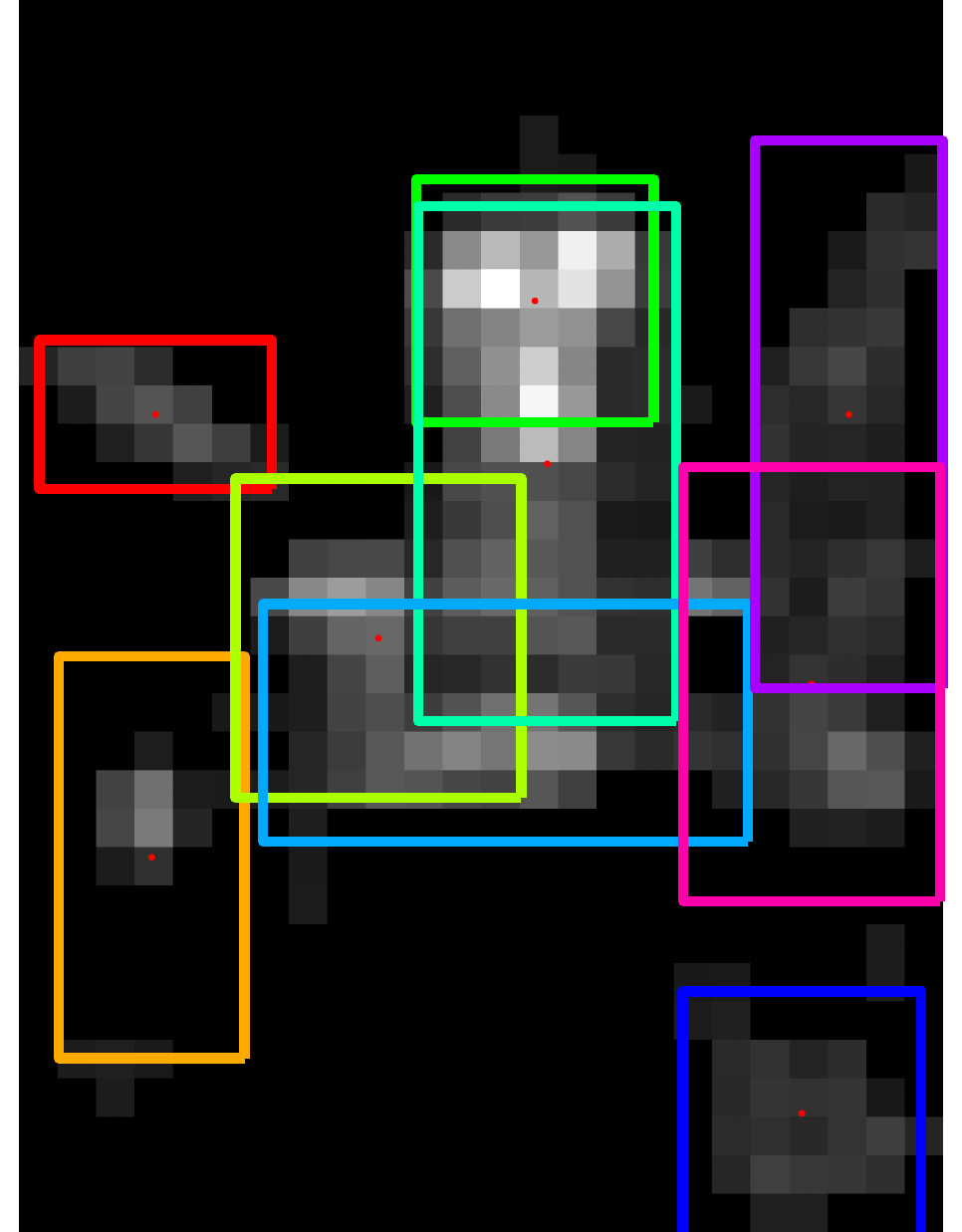} \\
image & $i=0,m=272$ & $i=2,m=29$ & $i=3,m=22$ & $i=5,m=17$ & $i=7,m=11$ & $i=14,m=9$ \\
\end{tabular}
\end{narrow}
\end{center}
\caption{Evolution of regions during EGM iterations on the feature saliency map of an image of \textit{Magdalen tower} from Oxford buildings dataset, shown on the left. Below each image we display the iteration $i$ and the number of regions $m$.}
\label{fig:egmfs}
\end{figure*}
%
We are given a 2d saliency map $S$, which can be either the feature saliency described in section~\ref{sec:crow} or the object saliency described in Section~\ref{sec:saliency}. We use an \emph{expanding Gaussian mixture} (EGM) model~\cite{AvKa12} to detect a number of salient rectangular regions. This is a variant of expectation-maximization (EM) that iteratively performs local averaging (E- and M-steps) interleaved with a selection process (P-step) similar to non-maximum suppression (NMS). In doing so, it dynamically estimates the number of regions.

The original algorithm applies to point sets and isotropic Gaussian components. Here we extend it to functions, considering that a saliency map is a function $S: P \to \real$. We use it to fit a number of components, each modeling a rectangular region in 2d coordinate space. We also extend it to a diagonal covariance model, so that a rectangle is modeled by an axis-aligned ellipse.

In particular, given 2d saliency map $S \in \real^{h \times w}$, we represent it as a set of Gaussian functions $s_i: \real^2 \to \real$ with
\begin{equation}
	s_i(\vx) \defn S_{p_i} \cN(\vx | p_i, \sigma I_2)
	\label{eq:sample}
\end{equation}
for $i \in [\ell]$, $\vx \in \real^2$ where $\cN$ is the normal density, $\ell = |P|$ is the number of positions and we represent $P$ as $\{p_1, \dots, p_\ell \}$. Here, $\sigma$ is a \emph{scale} parameter that determines how coarse or fine the region representation will be for the given saliency map. Similarly, we represent components as Gaussian functions $q_k: \real^2 \to \real$ with
\begin{equation}
	q_k(\vx) \defn \pi_k \cN(\vx | \mu_k, \Sigma_k)
	\label{eq:comp}
\end{equation}
for $k \in [m]$, $\vx \in \real^2$, where $m$ is the number of components and $\pi_k \in \real$, $\mu_k \in \real^2$ and $\Sigma_k \in \real^{2 \times 2}$ are the mixing coefficient, mean and diagonal covariance matrix respectively of component $k$. Means represent region centers, while the (inverse) eigenvalues of covariance matrices represent heights and widths. We initialize components as $q_k \gets s_k$ for $k \in [m]$, with $m \gets \ell$. In the \emph{expectation} (E)-step, we compute the \emph{responsibility}
\begin{equation}
	\gamma_{ik} \gets \frac{\inner{s_i}{q_k}}{\sum_{j \in [m]} \inner{s_i}{q_j}}
	\label{eq:gamma}
\end{equation}
of component $k \in [m]$ for sample $i\in [\ell]$, where $\inner{f}{g}$ is the $L^2$ inner product of square-integrable functions $f,g: \real^d \to \real$, computed in closed form for Gaussian functions~\cite{AvKa12}. In the \emph{maximization} (M)-step, we update parameters as
\begin{align}
	\pi_k & \gets \frac{\ell_k}{\ell}
	\label{eq:pi} \\
	\mu_k & \gets \frac{1}{\ell_k} \sum_{i=1}^n \gamma_{ik} p_i
	\label{eq:mu} \\
	\Sigma_k & \gets \frac{1}{\ell_k} \sum_{i=1}^n \gamma_{ik} \diag(p_i - \mu_k)^{\circ 2}
	\label{eq:sigma}
\end{align}
where $\ell_k \defn \sum_{i=1}^n \gamma_{ik}$ is the effective number of points assigned to component $k$ and $X^{\circ 2} \defn X \circ X$ is the Hadamard product power for a vector or matrix $X$.

Finally, in the \emph{purge} (P)-step, similarly to NMS, we process components in descending order of mixing coefficient and we decide whether to keep a component or not depending on its overlap with the collection of previously kept components. Overlap is measured by a generalized responsibility function similar to~\eqref{eq:gamma}, and again inner products are given in closed form~\cite{AvKa12}. This means that the number of components $m$ is potentially reducing at each iteration.

Figure~\ref{fig:egmfs} shows how regions are formed during EGM iterations, starting from one small region centered on each spatial position. We get 4 clean regions on the ground truth building, as well as 6 regions on background objects, which, although less salient, cannot be removed based on the feature saliency alone.

\subsection{Region pooling}
\label{sec:desc}
Given a rectangular region $R \in \cR$ of an image with feature saliency map $\hat{F}  \in \real^{h \times w}$, we associate to it \emph{feature saliency} $f \defn \mu_{\hat{F}}(R) \in \real$, where
\begin{equation}
	\mu_{\hat{F}}(R) \defn \frac{1}{|R|} \sum_{p \in R} \hat{F}_p
\label{eq:avg}
\end{equation}
is the average of 2d map $\hat{F}$ over $R$.

In addition, given the activation map $A \in \real^{h \times w \times c}$ of the same image, it is standard practice that a descriptor is obtained by pooling over $R$, for instance sum~\cite{BL15}, weighted sum~\cite{KMO15} or max~\cite{ARSM+14,TSJ15} pooling. We adopt the latter choice to extract descriptor $\vz \defn m_A(R) \in \real^c$, where
\begin{equation}
	m_A(R) \defn \max_{q \in R} A_{q \bdot}
\label{eq:max}
\end{equation}
is the maximum of 3d tensor $A$ over $R$ along the spatial dimensions.
This has been the basis of fine-tuning in~\cite{RTC16,GARL16}.

A particular set of regions, uniformly sampled on a grid at different scales, is referred to as \emph{regional maximum activation of convolutions} (R-MAC)~\cite{TSJ15}. Global description, referred to as MAC, is a special case where there is a single region $R = P$. In contrast, we detect a set of regions based on saliency maps in this work.

Finally, we follow~\cite{RTC16} in performing supervised whitening of the descriptors by simultaneous diagonalization~\cite{MiMa07}. In particular, given vector $\vz \in \real^c$, we $\ell^2$-normalize, center,
 whiten, PCA-project and renormalize to generate the \emph{region descriptor} $\vv \defn w(\vz) \in \real^d$ for region $R$. Function $w: \real^c \to \real^d$ represents this pipeline entirely.

\subsection{Graph construction}
\label{sec:graph}
Given an image dataset, we assume here a set of regions $\{ R_1, \dots, R_n \}$ are detected from the saliency maps (Section~\ref{sec:egm}), a \emph{feature saliency} vector $\vf \defn (f_1, \dots, f_n) \in \real^n$ is computed with the corresponding average saliency per region in (\ref{eq:avg}), and a set of descriptors $V \defn \{ \vv_1, \dots, \vv_n \} \subset \real^d$ are extracted from the activation maps, whitened and normalized per region (Section~\ref{sec:desc}).

Based on the above information, we construct a $k$-NN graph on those regions in order to compute a global \emph{centrality} score per region as discussed in Section~\ref{sec:katz}, which enables us to form an \emph{object saliency} map on a new image, described in Section~\ref{sec:saliency}.
Approximate techniques for $k$-NN graph construction~\cite{DCL11} can be used to handle large-scale databases.

We construct a weighted undirected graph having the set of descriptors $V$ as vertices. Following~\cite{ITA+16}, the edge weights are defined according to \emph{mutual $k$-nearest neighbors} (NN) in the descriptor space. In particular, given descriptors $\vv, \vu \in \real^d$, we measure their \emph{similarity} by $s(\vv, \vu) = (\vv^\T \vu)^\beta$, where exponent $\beta > 0$ is a parameter. We define the sparse symmetric nonnegative \emph{adjacency matrix} $W \in \real^{n \times n}$ with elements $w_{ij}$ being $s(\vv_i, \vv_j)$ if $\vv_i, \vv_j$ are mutual $k$-NN in $V$ and zero otherwise.

We define the $n \times n$ \emph{degree matrix} $D \defn \diag(W \vone)$ where $\vone \in \real^n$ is the all-ones vector, and the \emph{symmetrically normalized adjacency matrix}
\begin{equation}
	\cW \defn D^{-1/2} W D^{-1/2},
\label{eq:adj}
\end{equation}
with the convention $0/0 = 0$. Following~\cite{ITA+16,IAT+17}, we define the $n \times n$ matrices $L_\alpha \defn (D - \alpha W) / (1 - \alpha)$ and
\begin{equation}
	\cL_\alpha \defn D^{-1/2} L_\alpha D^{-1/2} = (I - \alpha \cW) / (1 - \alpha),
\label{eq:lap}
\end{equation}
where $\alpha \in [0,1)$. Both are positive-definite~\cite{ITA+16,IAT+17}.

\subsection{Graph centrality}
\label{sec:katz}
With the above definitions in place, the objective is to compute a vector $\vg \in \real^n$ where each element $g_i$ represents the significance of vertex $\vv_i$ in the graph, for $i \in [n]$. We define this \emph{centrality vector} as the solution $\vg^* \in \real^n$ of the linear system
\begin{equation}
	\cL_\alpha \vg = \vone.
\label{eq:katz}
\end{equation}
As in~\cite{ITA+16}, we solve this system by the \emph{conjugate gradients} (CG)~\cite{NoWr06} method. Any method would be equally appropriate because this is computed just once offline.

The solution $\vg^*$ is a \emph{graph centrality} measure~\cite{Newm10}, and in particular, \emph{Katz centrality}~\cite{Katz53}. Centrality is a global measure of significance of vertices in a graph, and PageRank~\cite{PBM+99} is maybe the most well-known. In fact, Katz centrality was introduced as such a global measure before being adapted by \emph{boundary condition} $\vy$ to measure relevance to individual vertices by Hubbell~\cite{Hubb65}. This work has a long history before being rediscovered \eg by~\cite{PBM+99,ZWG+03}, as summarized in the study of \emph{spectral ranking}~\cite{Vign09}.

\begin{figure}
\centering
\includegraphics[width=.48\textwidth]{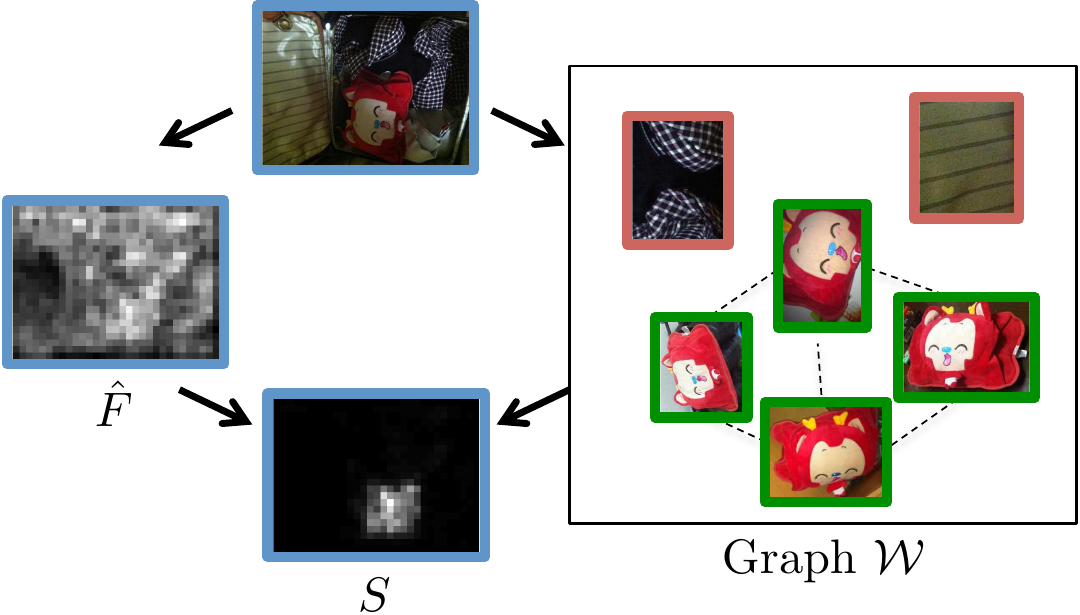}
\caption{Computing the \emph{object saliency} map $S$ of an image from Instre dataset (top), as defined in~\eqref{eq:obj}. For each patch, its neighbors in the graph (right) are found. Common patterns with high centrality in green outline, outliers with low centrality in red. $S$ (bottom) then focuses on patches similar to common patterns and combines with feature saliency $\hat{F}$ (left).
}
\label{fig:osMain}
\vspace{-10pt}
\end{figure}

\subsection{Saliency map construction}
\label{sec:saliency}
Given the region descriptor set $V$, the region saliency vector $\vf$ and the associated centrality vector $\vg^*$ of an entire dataset, the problem is to construct a new saliency map $S \in \real^{h \times w}$ for an image in the dataset. The image is represented by its activation map $A \in \real^{h \times w \times c}$. Since this saliency is based on regions or patterns appearing frequently in the dataset, which are commonly associated to repeating objects, we call it \emph{object saliency} (\OS).

We compute $S$ by a sliding window iteration over each position $p \in P$. The saliency value $S_p$ at $p$ is found as a linear combination of the centrality values of the nearest neighbors in $V$ of a patch centered at $p$. In particular, we consider a square patch $R_p$ of side $a$ centered at $p$. We compute the vector $\vu_p \defn w(m_A(R_p)) \in \real^d$ by max-pooling over $R_p$, whitening and normalizing as discussed in Section~\ref{sec:desc}. If $N_p$ is the set of indices of the $k$-NN of $\vu_p$ in $V$, we compute $S_p$ as
\begin{equation}
	S_p \defn \hat{F}_p^\queryPow \sum_{i \in N_p} s(\vv_i, \vu_p) f_i^\dbPow g_i^*.
\label{eq:obj}
\end{equation}
That is, each neighboring region descriptor $\vv_i$ is weighted by its similarity to patch descriptor $\vu_p$, its feature saliency $f_i$ and its centrality $g_i^*$, while the entire sum is scaled by the feature saliency $\hat{F}_p$ at the current position $p$ of the image being considered. Exponents $\Theta$ and $\theta$ control the relative importance of feature saliency of the current image and neighbors, respectively, compared to centrality.
The object saliency computation is illustrated in Figure~\ref{fig:osMain}. Looking at the input image and is feature saliency map $\hat{F}$ alone, it is not evident which is the object of interest and which is clutter. This is only found by discovering other instances of the same object in the dataset, as represented by the graph.

\subsection{Representation}
\label{sec:repr}
The object saliency map $S$ highlights patterns that appear frequently in the dataset, with the background clutter removed. It is only natural then to apply the same method described in Section~\ref{sec:egm} to this map in order to detect a small number of regions per image. Unlike the regions detected from the feature saliency map $\hat{F}$, these new regions are more likely to appear in a new image. For the purpose of evaluation, we investigate both saliency maps.

For each region $R$ detected from a saliency map ($\hat{F}$ or $S$) in a dataset image with activation map $A$, we apply max pooling and $\ell^2$-normalization.
All descriptors are then summed and the resulting descriptor is whitened with $w: \real^c \to \real^d$ as described in Section~\ref{sec:desc}. The difference here is that we apply whitening on the aggregated vector and not separately per region. This is the same representation as R-MAC evaluated in~\cite{RTC16} and both yield a global image representation in $\real^d$, but here the regions are detected in the saliency map rather than being uniformly distributed.

Pooling based on saliency is in fact the idea explored in CroW~\cite{KMO15}, but here we follow the nonlinear two-level pooling of R-MAC (max followed by sum) rather than the one-level sum of CroW. This is more powerful and has also been the basis of fine-tuning in~\cite{GARL16}.

\section{Experiments}
\label{sec:exp}
We apply the proposed representation on image retrieval.
In particular, we have two variants of our method that both use the region detection described in Section~\ref{sec:egm}.
The saliency map which the detection is performed on is different in each case.
\FSegm uses the feature saliency map described in Section~\ref{sec:crow}, and
\OSegm uses the object saliency map described in Section~\ref{sec:egm}.
The former is image specific, while the latter both image and database specific.

\subsection{Experimental setup}
\medskip
\head{Test sets.}
We evaluate on Oxford Buildings~\cite{PCISZ07} and the more recently introduced Instre~\cite{WJ15} dataset.
Instre contains around 27k images of small objects in cluttered scenes while objects appear with different variations, such as rotation, occlusion and scale changes, making it a challenging case.
We use the evaluation protocol introduced in~\cite{ITA+16} for Instre.
We add 100k distractors from Flickr~\cite{PCISZ07} to Oxford5k to perform experiments at larger scale. We refer to it as Oxford105k.
Search performance in all datasets is measured with mAP.

\medskip
\head{Image Representation.}
We represent each image by global image representation as described in Section~\ref{sec:repr}.
This reduces image similarity to cosine, which is common practice~\cite{TSJ15}.
Feature extraction is performed with the VGG network~\cite{SZ14} that is fine-tuned specifically for image retrieval~\cite{RTC16}.
Supervised whitening~\cite{RTC16} is used for post-processing.
The same network is additionally used to compare against two baselines.
First, MAC global descriptor, which is obtained by global max pooling and the descriptor that the network is directly optimized for~\cite{RTC16}.
Second, the baseline approach (\textit{Uniform}), which refers to regional max pooling for regions that are uniformly sampled at 3 scales, as in R-MAC~\cite{TSJ15}.
Our variants are different in that regions are detected from salient and repeating objects, while aggregation and whitening is identical.
Detection is applied to dataset images only, while we use the provided bounding boxes on the query side.

\medskip
\head{Implementation Details.}
To simplify region detection, each saliency map is masked above threshold $\cutoff$ and element-wise raised to exponent $\salPow$ before detection,
which removes the weakest regions and increases the contrast between foreground and background objects.
We set $\salPow=1$, $\cutoff=0.2$ and scale parameter $\scale=1$ before any parameter tuning is performed.
We determine \OS parameters $\queryPow$, $\dbPow$ in~\eqref{eq:obj} by visual inspection of \OS and set $\queryPow=2$, $\dbPow=3$ throughout our experiments.
We perform our experiments on a 16-core Intel Xeon 2.00GHz CPU. It takes 36s to create the graph on Instre, while centrality computation takes negligible amount of time.
It takes 0.02s for FS computation and detection per image, while 0.23s in the case of OS.

\subsection{Parameter tuning}
\label{sec:fsFinetune}

In this section, we show the impact of \FSegm and \OSegm detection parameters on the retrieval performance.
We tune the parameters on Oxford5k when using diffusion~\cite{ITA+16}.
The remaining experiments evaluate the proposed representation with the chosen parameters on Instre and Oxford105k as well.

\begin{figure}
\centering
\input{figs/data/sample}
\small
\begin{tabular}{c}
\extfig{fsPow}{
\begin{tikzpicture}
\begin{axis}[%
	width=0.9\linewidth,
	height=0.4\linewidth,
	xlabel={Exponent $\salPow$},
	ylabel={mAP},
	legend cell align={left},
	legend pos=south east,
    legend style={cells={anchor=east}, font =\scriptsize, fill opacity=0.8, row sep=-2.5pt},
    xmax = 6,
    xmin = 1,
    ymin = 83,
   	xtick={1,2,...,6},
    grid=both,
]
	\addplot[color=blue,     solid, mark=*,  mark size=1.5, line width=1.0] table[x=pow, y expr={\thisrow{oxford5k}}] \osPow;\leg{\OSegm};
	\addplot[color=red,     solid, mark=*,  mark size=1.5, line width=1.0] table[x=pow, y expr={\thisrow{oxford}}] \fsPow;\leg{\FSegm};

\end{axis}
\end{tikzpicture}
}

\end{tabular}
\caption{mAP on Oxford5k versus saliency exponent $\salPow$
for \FSegm and \OSegm.
\vspace{-10pt}
}
\label{fig:fsPow}
\end{figure}

\begin{figure}
\centering
\input{figs/data/sample}
\small
\begin{tabular}{c}
\extfig{fsCutoff}{
\begin{tikzpicture}
\begin{axis}[%
	width=0.9\linewidth,
	height=0.4\linewidth,
	xlabel={Threshold $\cutoff$},
	ylabel={mAP},
	legend cell align={left},
	legend pos=south east,
    legend style={cells={anchor=east}, font =\scriptsize, fill opacity=0.8, row sep=-2.5pt},
    xmax = 0.6,
    xmin = 0,
   	xtick={0.1,0.2,...,0.6},
    grid=both,
]
	\addplot[color=red,     solid, mark=*,  mark size=1.5, line width=1.0] table[x=cutoff, y expr={\thisrow{oxford5k}}] \fsCutoff;\leg{\FSegm};
	\addplot[color=blue,     solid, mark=*,  mark size=1.5, line width=1.0] table[x=cutoff, y expr={\thisrow{oxford5k}}] \osCutoff;\leg{\OSegm};

\end{axis}
\end{tikzpicture}
}

\end{tabular}
\caption{mAP on Oxford5k versus threshold $\cutoff$
for \FSegm and \OSegm.
\vspace{-10pt}
}
\label{fig:fsCutoff}
\end{figure}

\begin{figure}
\centering
\input{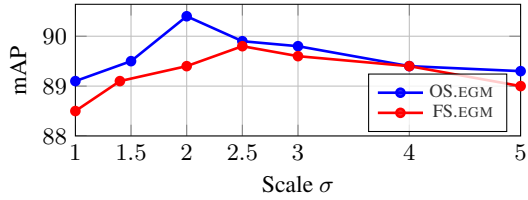}
\small
\begin{tabular}{c}
\extfig{fsScale}{
\begin{tikzpicture}
\begin{axis}[%
	width=0.9\linewidth,
	height=0.4\linewidth,
	xlabel={Scale $\scale$},
	ylabel={mAP},
	legend cell align={left},
	legend pos=south east,
    legend style={cells={anchor=east}, font =\scriptsize, fill opacity=0.8, row sep=-2.5pt},
    xmax = 5,
    xmin = 1,
    ymin = 88,
   	xtick={1,1.5,2,2.5,3,4,5},
    grid=both,
]
	\addplot[color=blue,     solid, mark=*,  mark size=1.5, line width=1.0] table[x=scale, y expr={\thisrow{oxford5k}}] \osScale;\leg{\OSegm};
	\addplot[color=red,     solid, mark=*,  mark size=1.5, line width=1.0] table[x=scale, y expr={\thisrow{oxford5k}}] \fsScale;\leg{\FSegm};

\end{axis}
\end{tikzpicture}
}

\end{tabular}
\caption{mAP on Oxford5k versus EGM scale parameter $\scale$ for \FSegm and \OSegm.
\vspace{-10pt}
}
\label{fig:fsScale}
\end{figure}

\medskip
\head{Feature saliency detection} is evaluated first by \FSegm, while we do not compute object saliency and \OSegm yet.
Figure~\ref{fig:fsPow} shows the effect of $\salPow$, which controls the contrast of the the saliency map.
We observe that large $\salPow$ is needed to remove as much clutter as possible from the noisy \FS activations.
We set $\salPow = 5$ for the rest of our experiments.
Figure~\ref{fig:fsCutoff} shows the effect of threshold $\cutoff$, which is another selectivity parameter. We set $\cutoff = 0.4$.
Scale $\scale$ is used during EGM sampling as explained in Section~\ref{sec:egm}. Its impact in performance is shown in Figure~\ref{fig:fsScale}.
Setting $\scale = 2.5$ results in good performance and regions that are large enough for \FSegm.

\head{Object saliency detection} is then evaluated once the feature saliency parameters are fixed,
and EGM detection is applied on the new saliency map.
We observe that \OS behaves quite differently to \FS, because foreground objects are much cleaner.
The impact of parameters $\scale$ and $\salPow$ is shown in Figures~\ref{fig:fsPow} and~\ref{fig:fsScale} respectively. It is remarkable that a much lower exponent is needed in this case.
We choose $\salPow = 2$ and $\scale =2$.
Finally, we fix $\cutoff = 0$ for \OS, as the saliency maps obtained with \OS are exactly zero at background regions.
The effect is shown in Figure~\ref{fig:fsCutoff}.

\subsection{Evaluation of saliency maps}

\begin{figure}
\centering
\input{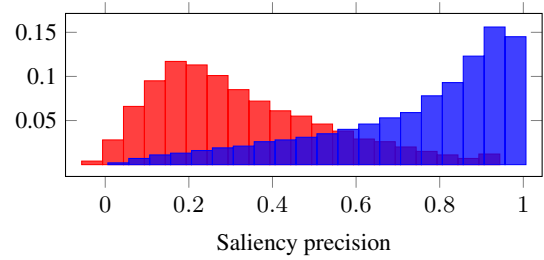}
\small
\begin{tabular}{c}
\extfig{regDet}{
\begin{tikzpicture}
\begin{axis}[%
	width=0.95\linewidth,
	height=0.45\linewidth,
	ybar,
	legend pos=north west,
    ytick={0.05,0.1,0.15},
    yticklabels={0.05,0.1,0.15},
    xlabel={Saliency precision},
 	 y label style={at={(axis description cs:-0.1,.5)}},
 	 x label style={at={(axis description cs:.5,-.3)}},
]
\addplot[color=red, fill = red,fill opacity=0.75,inner ysep=0.5pt, bar width = 0.05] table[x=binEdges, y expr={\thisrow{FS}}]  \salCorrectness;
\addplot[color=blue, fill = blue,fill opacity=0.75,inner ysep=0.5pt, bar width = 0.05] table[x=binEdges, y expr={\thisrow{OS}}]  \salCorrectness;
\end{axis}
\end{tikzpicture}
}
\end{tabular}
\caption{
Histogram of saliency precision for {\color{red}\FS} and {\color{blue}\OS} maps measured on all images of Instre.
\vspace{-10pt}
}
\label{fig:regDet}
\end{figure}

\begin{figure}
\setlength{\fboxsep}{0pt}%
\setlength{\fboxrule}{1.5pt}%
\newcommand{\showIm}[3]{\includegraphics[width=1.6cm,height=1.6cm]{figs/examples/#1_q_#2_#3.jpg}}
\newcommand{\qIm }[2]{\includegraphics[width=1.6cm,height=1.6cm]{figs/examples/{#2}_q_#1_bbx.jpg}}
\newcommand{\pr}[2]{\footnotesize{{\color{red}#1}{\scriptsize$\rightarrow$}{\color{blue}#2}}}

\newcommand{\posEx}[1]{\footnotesize{\color{blue}#1}}
\newcommand{\negEx}[1]{\footnotesize{\color{red}#1}}

\begin{center} \footnotesize
   \begin{tabular}
   {*{5}{@{\sssp}c@{\msp}}}
\includegraphics[height=1.8cm]{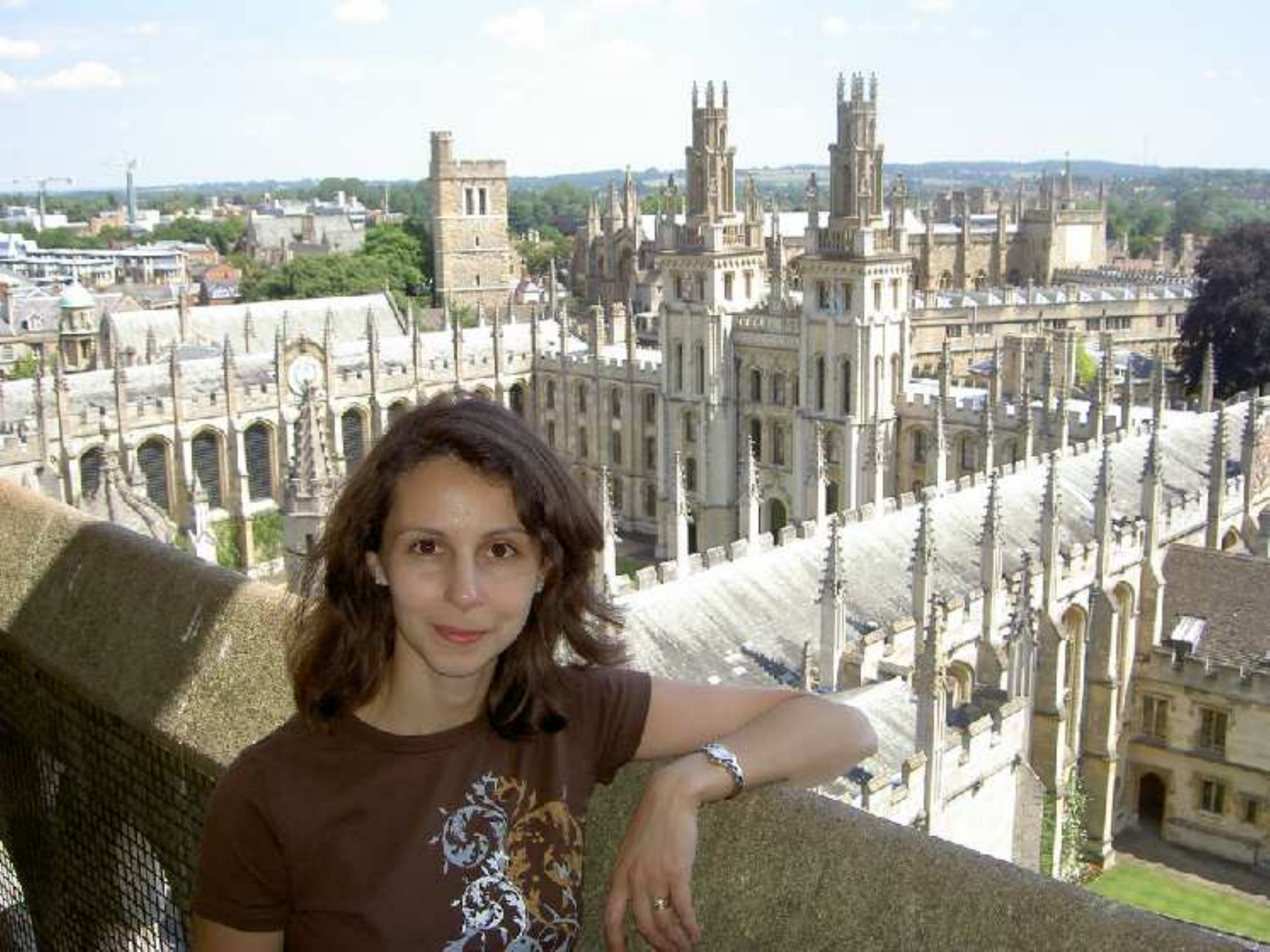} &
\includegraphics[height=1.8cm]{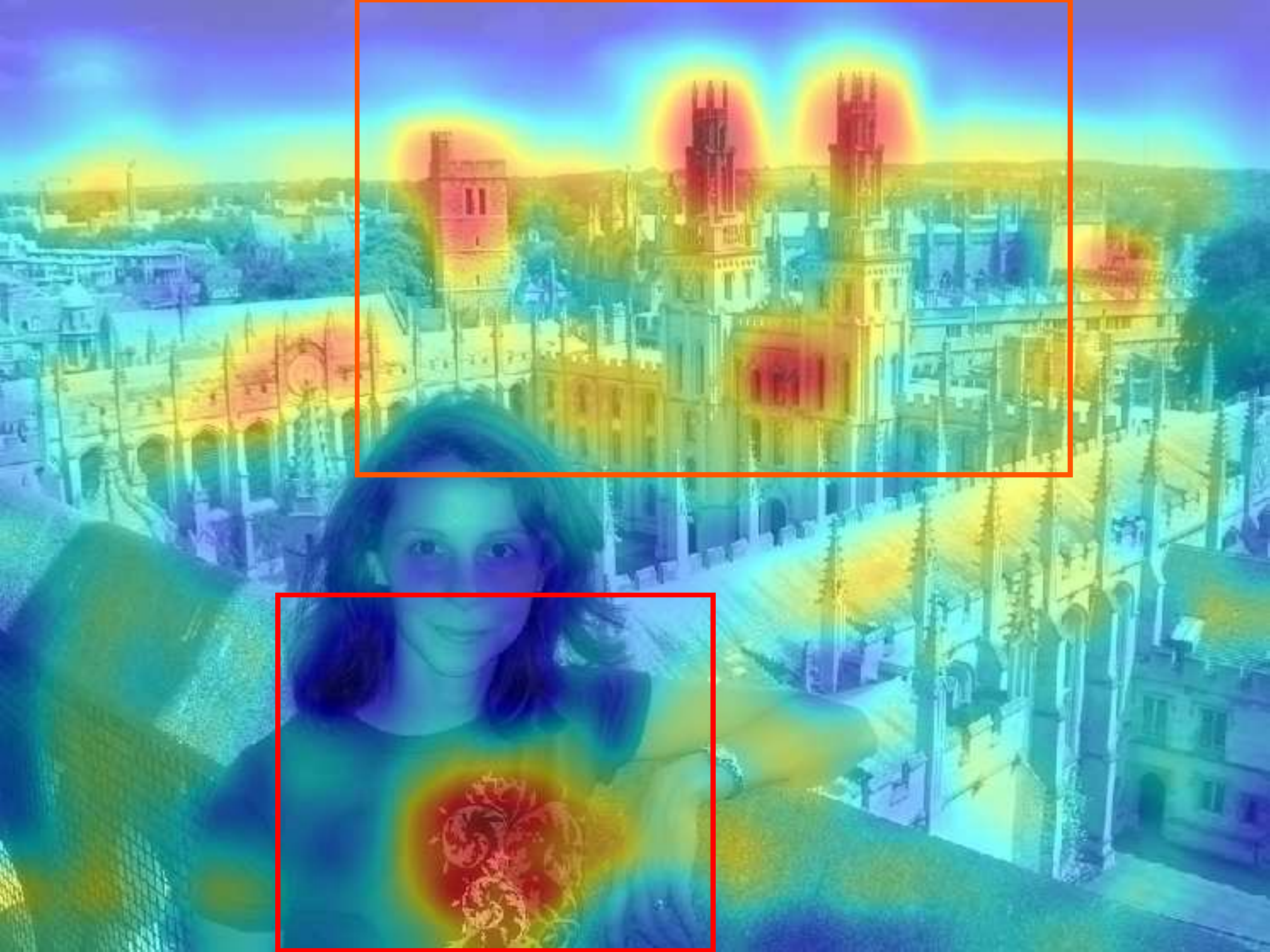} &
\includegraphics[height=1.8cm]{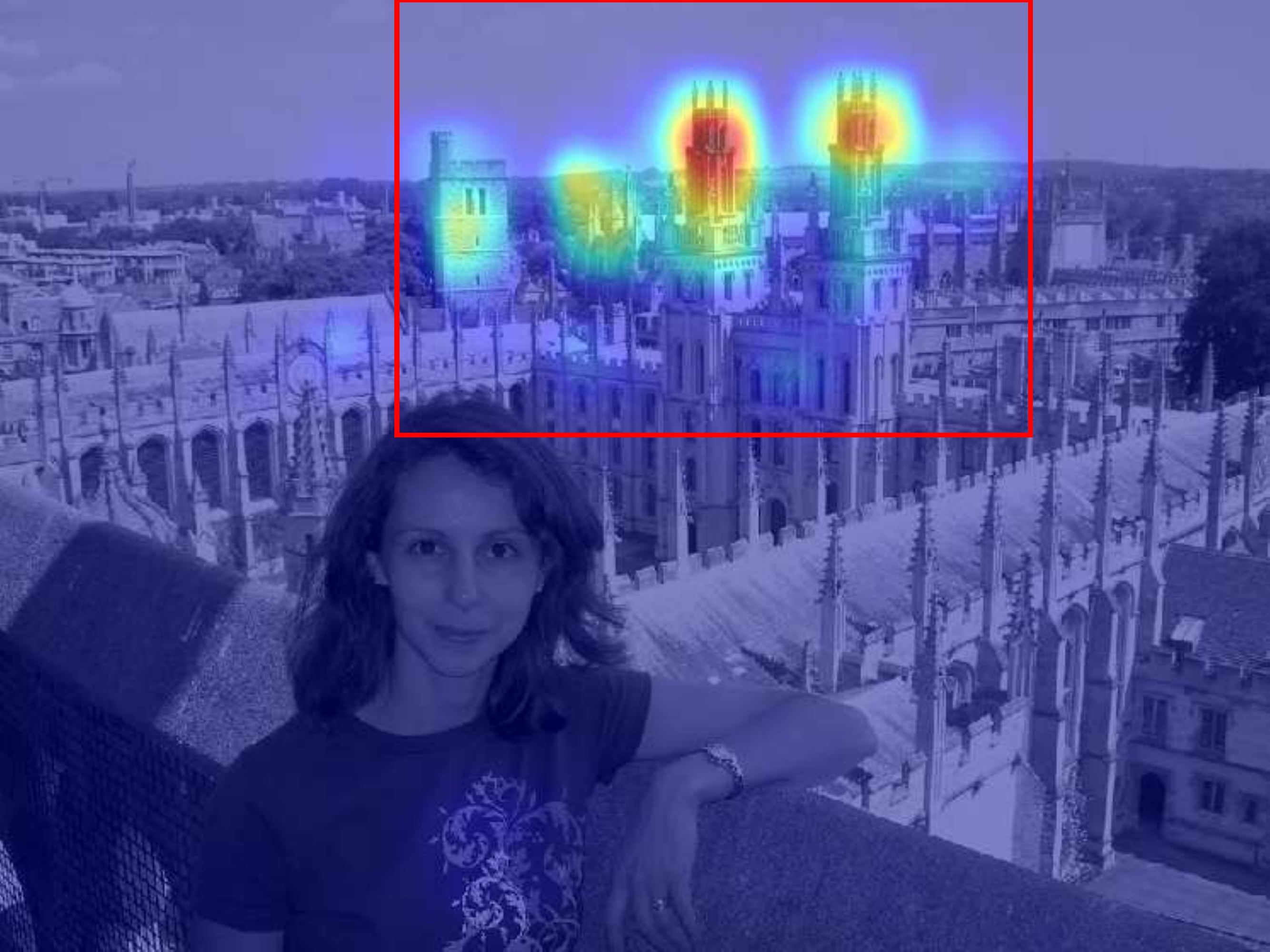} \\[2pt]
&&&&\\[-8pt]
\includegraphics[height=1.8cm]{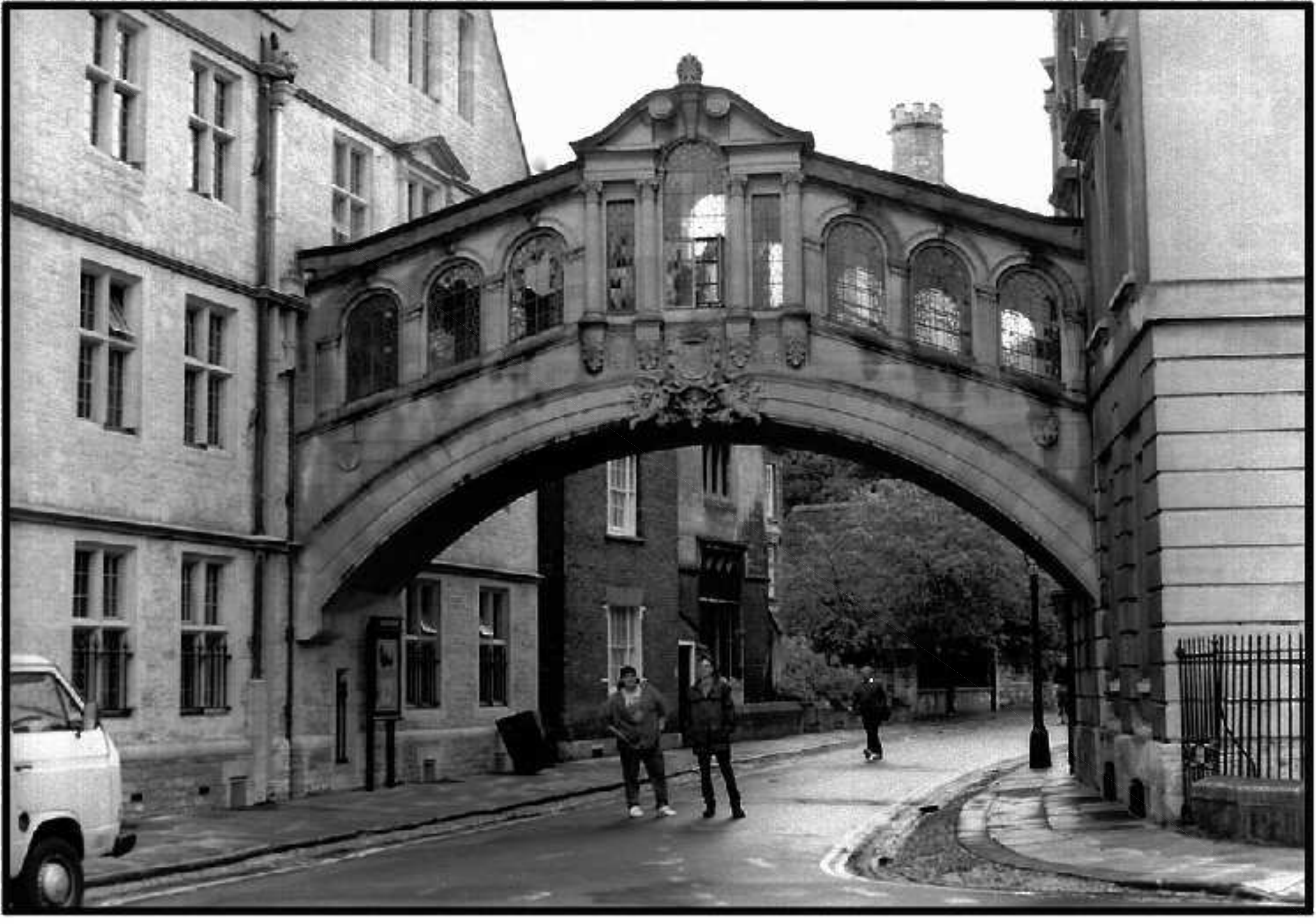} &
\includegraphics[height=1.8cm]{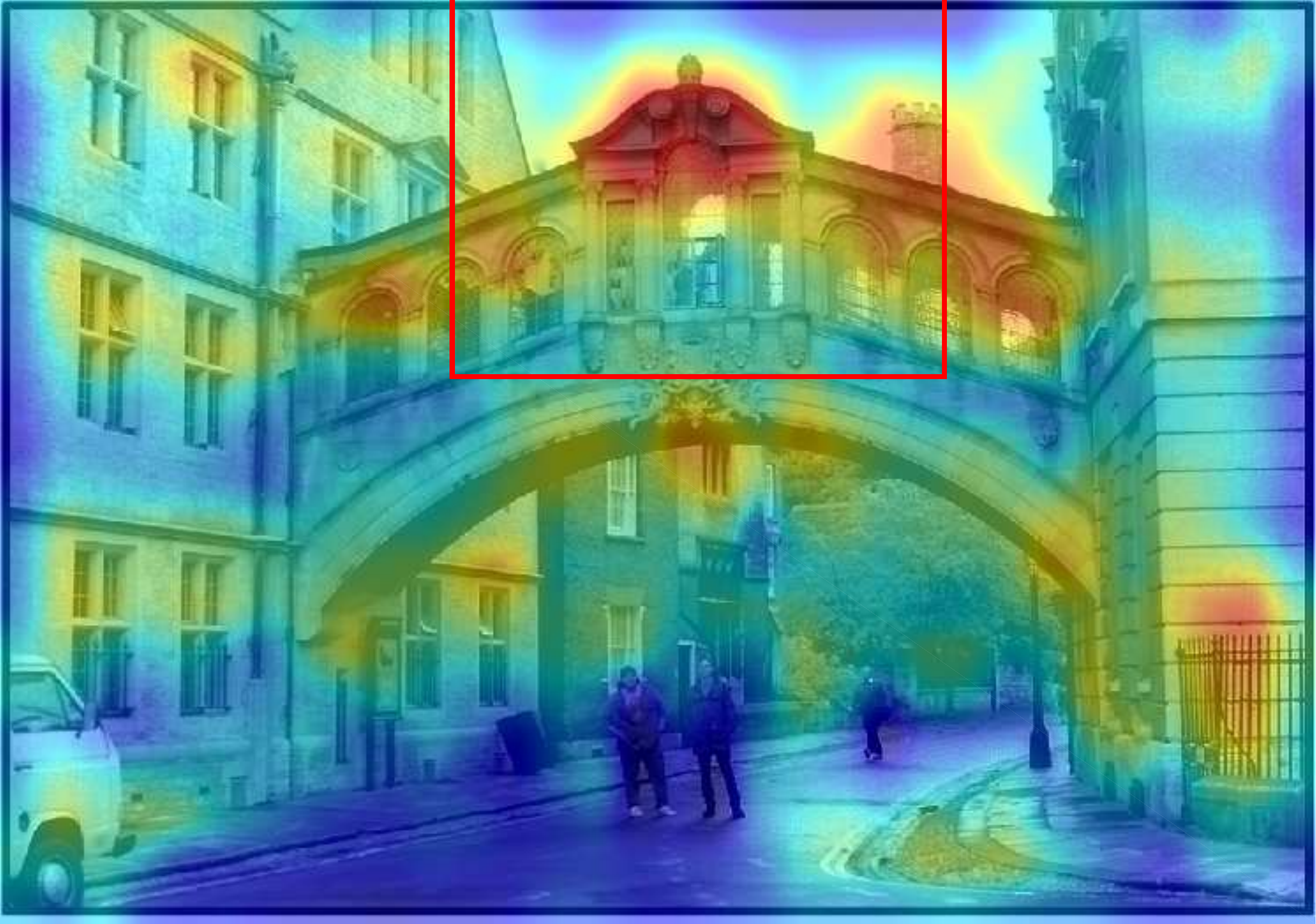} &
\includegraphics[height=1.8cm]{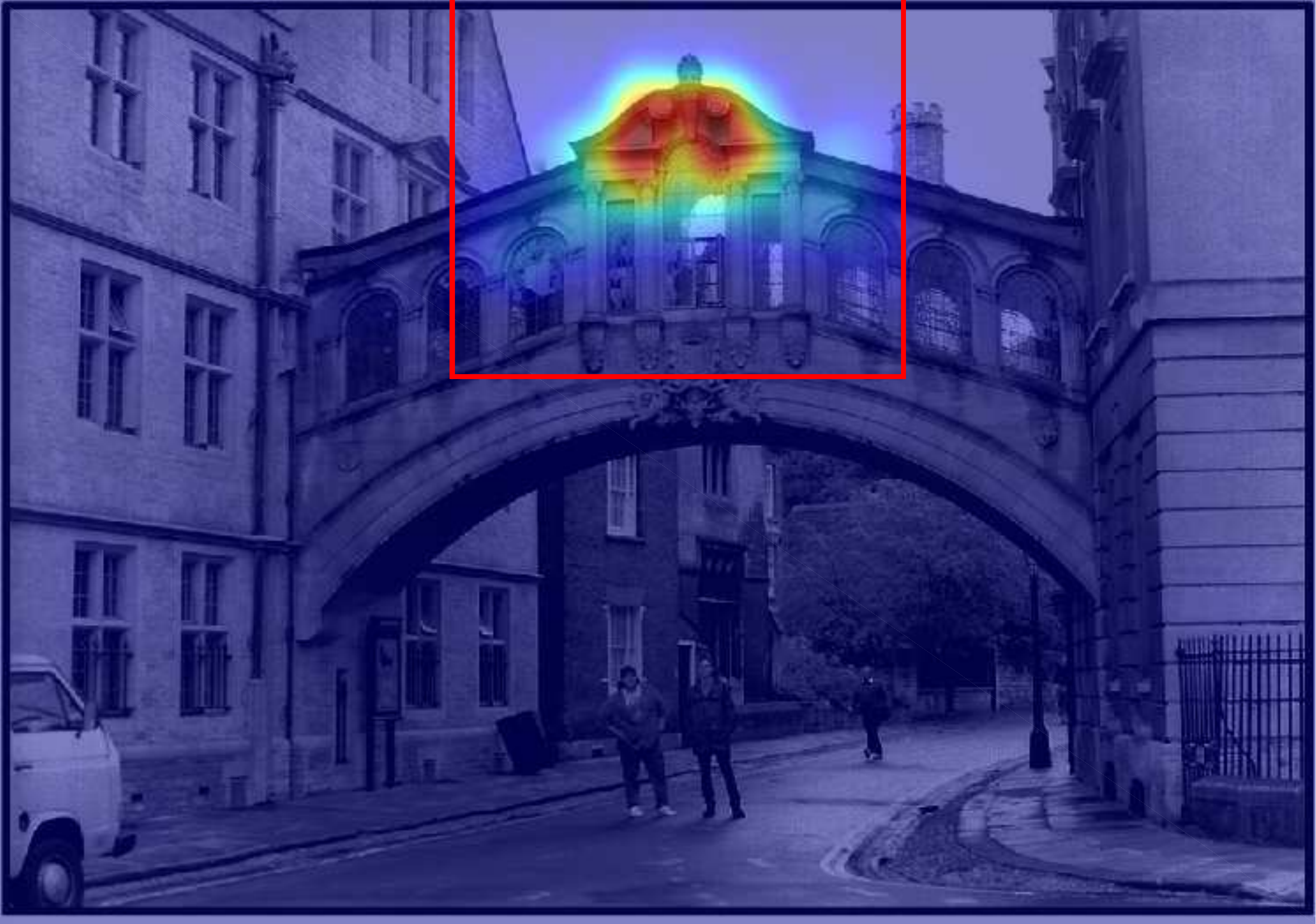} \\[2pt]
&&&&\\[-8pt]
\includegraphics[height=1.8cm]{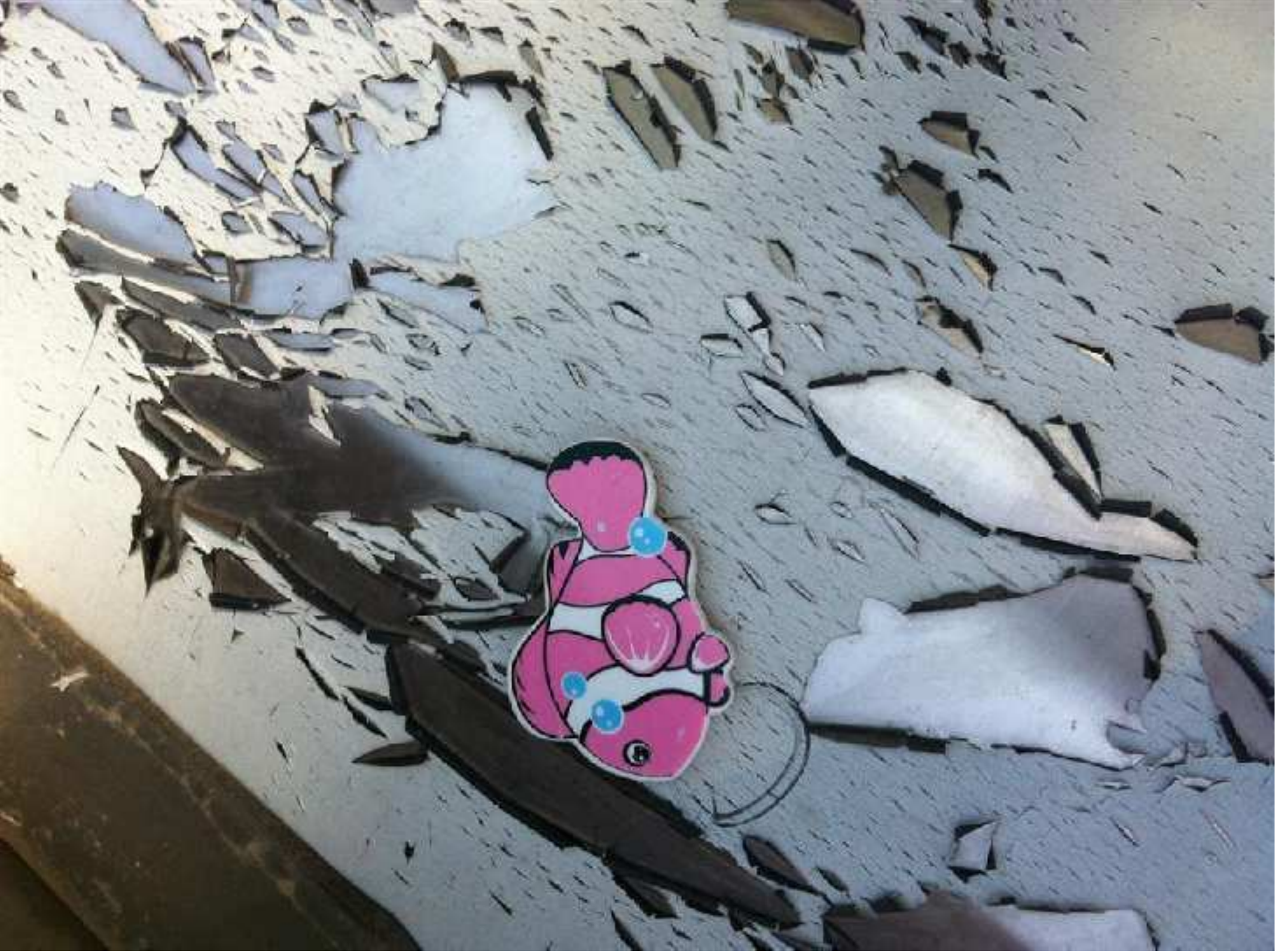} &
\includegraphics[height=1.8cm]{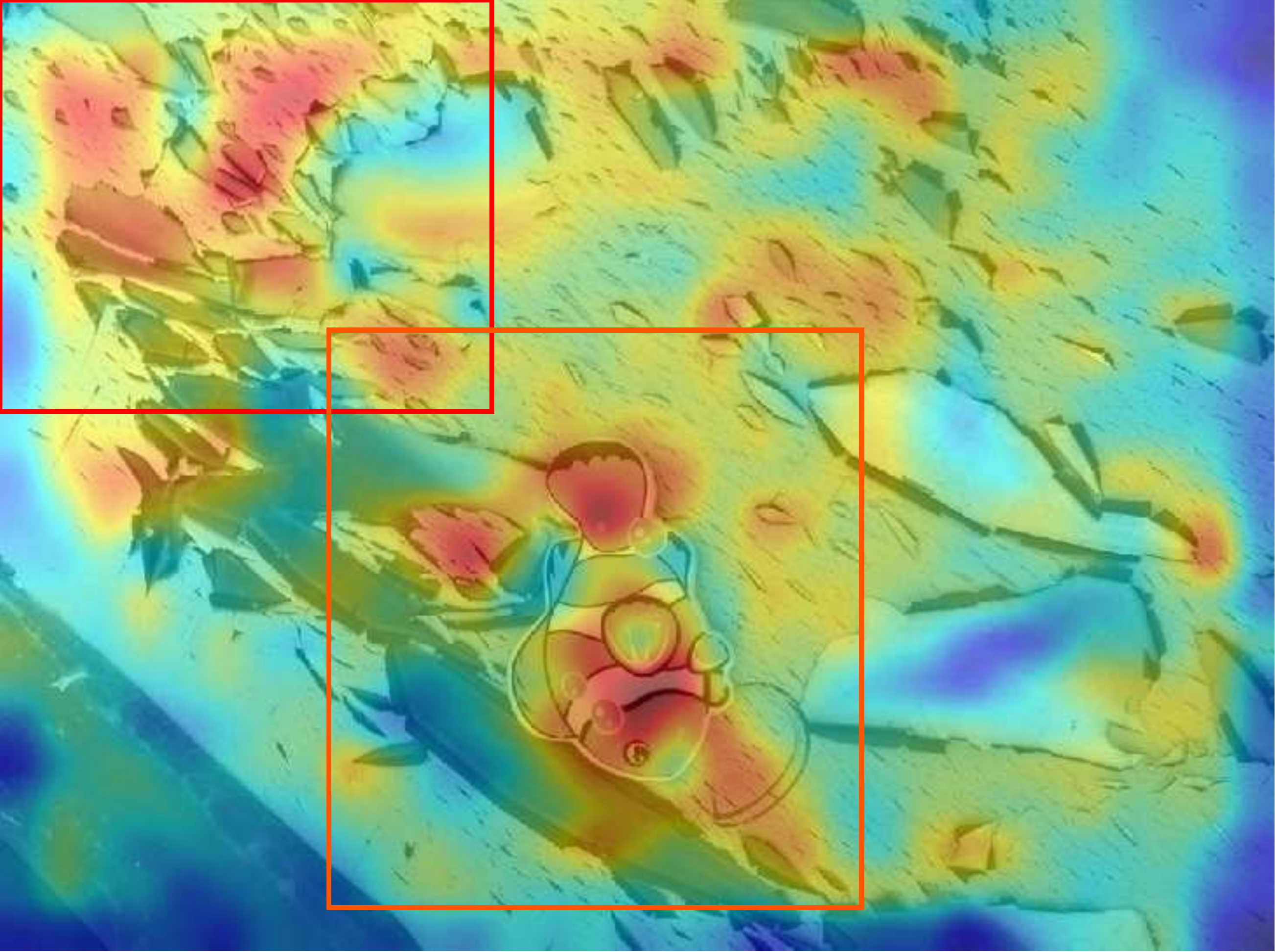} &
\includegraphics[height=1.8cm]{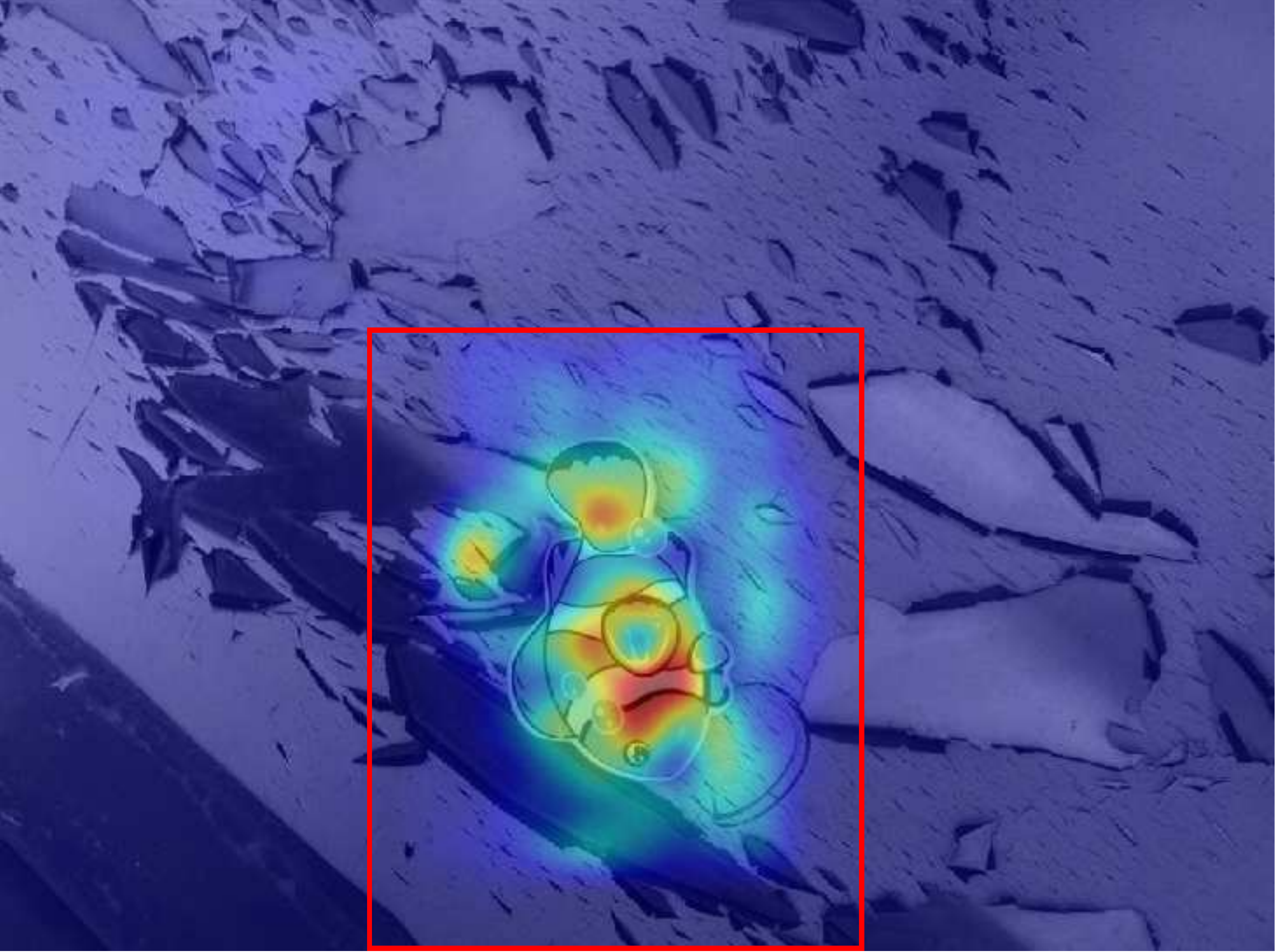} \\[2pt]
&&&&\\[-8pt]
\includegraphics[height=1.8cm]{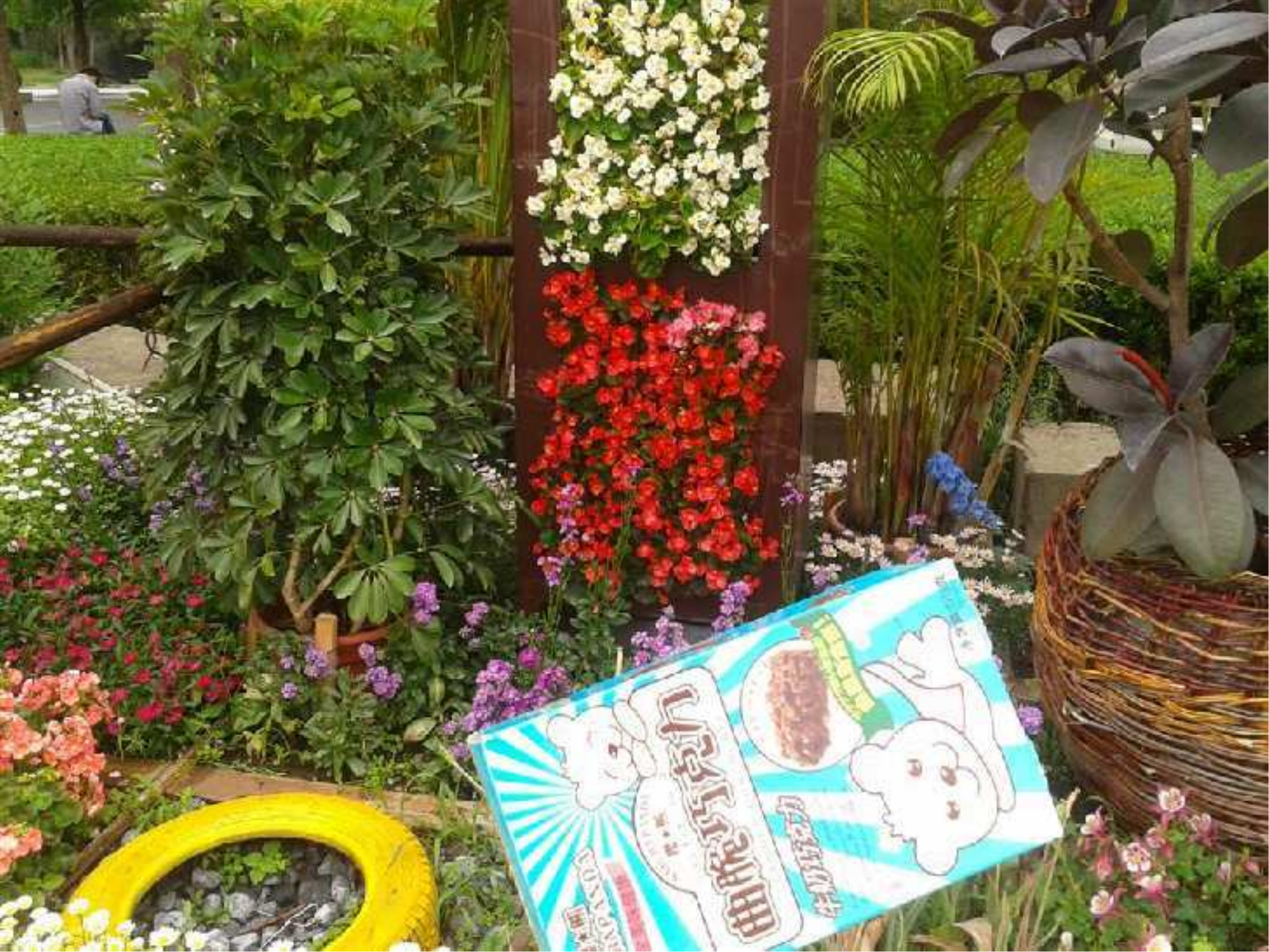} &
\includegraphics[height=1.8cm]{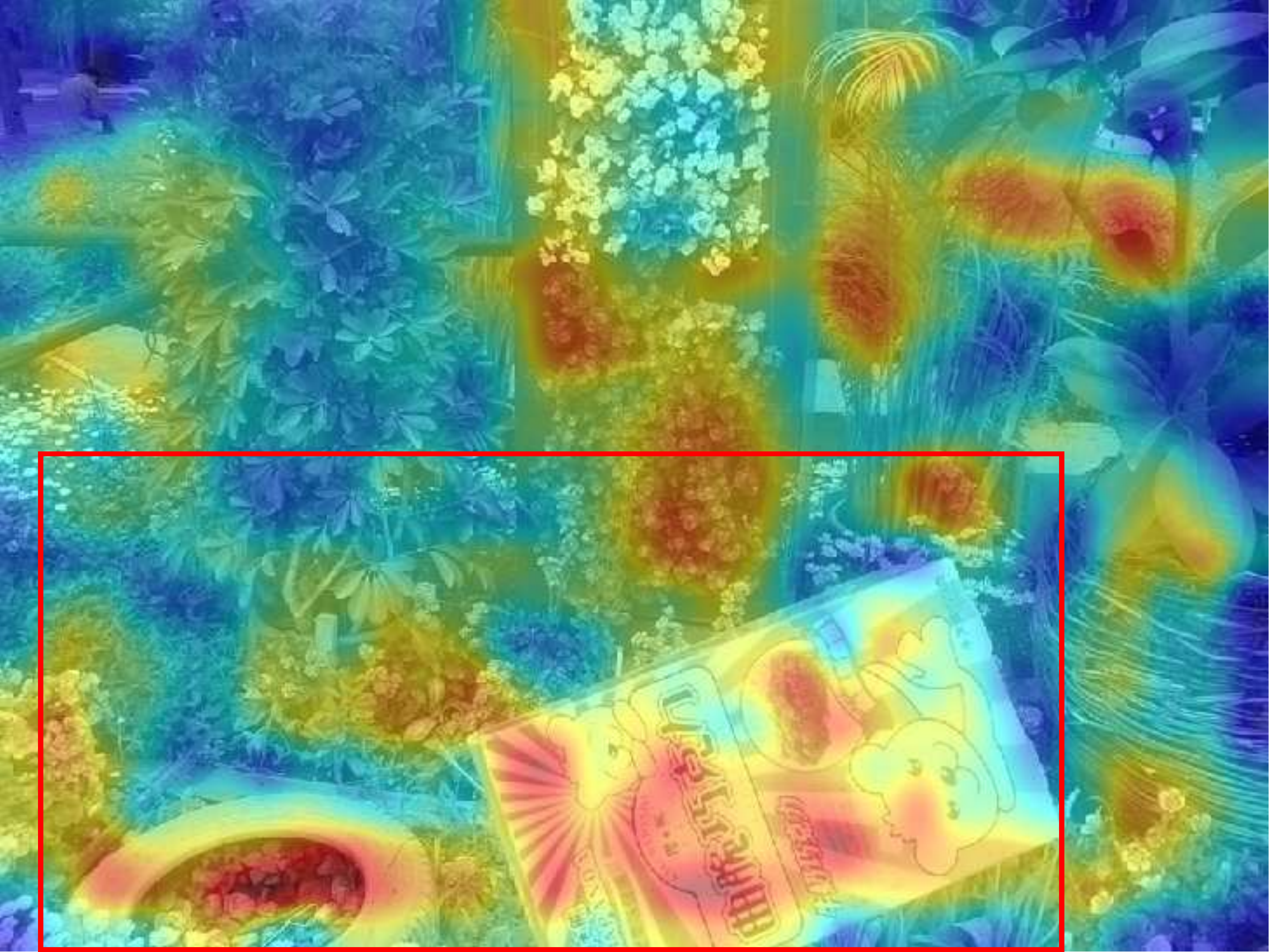} &
\includegraphics[height=1.8cm]{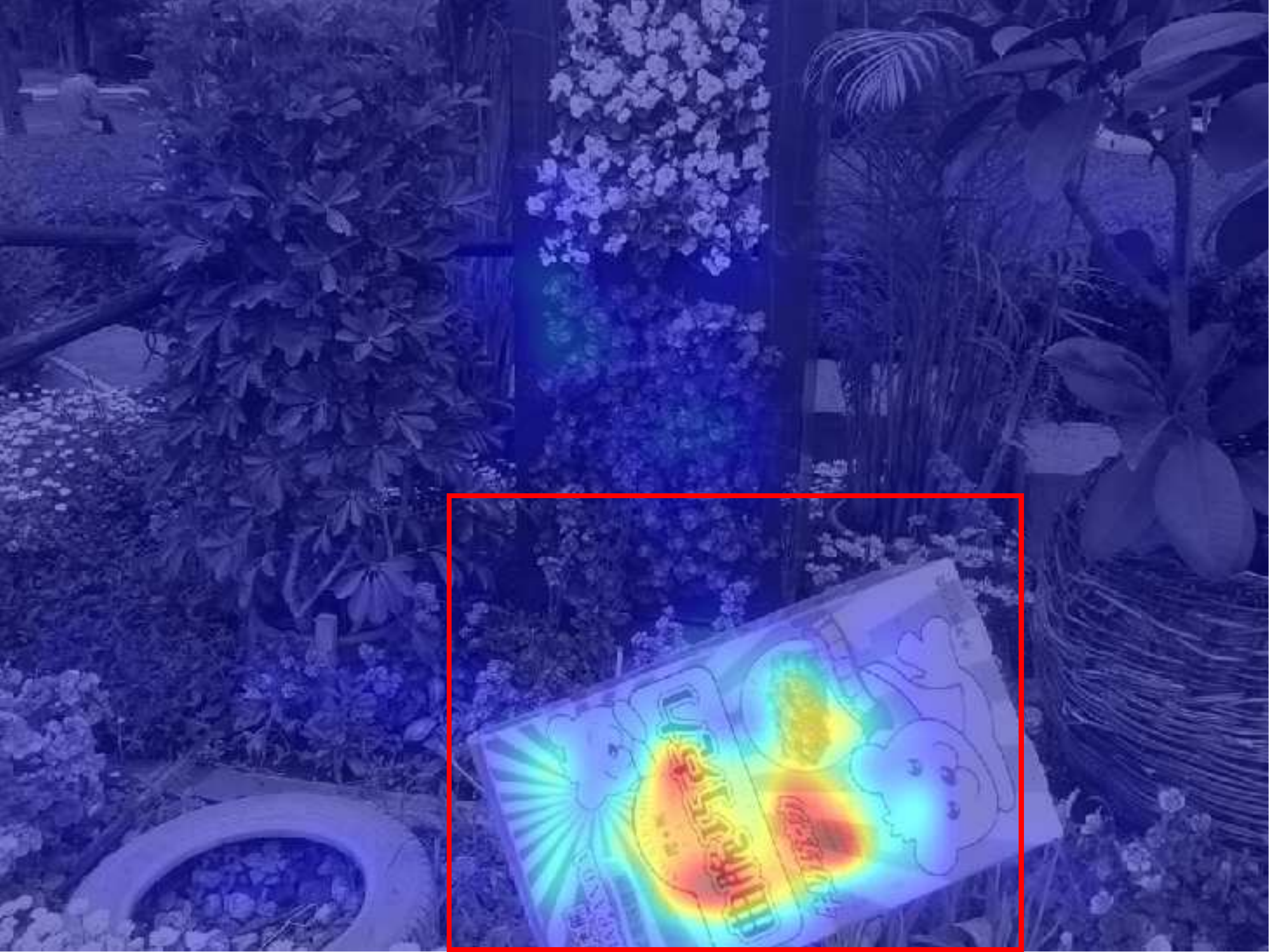} \\[2pt]
&&&&\\[-8pt]
\includegraphics[height=1.8cm]{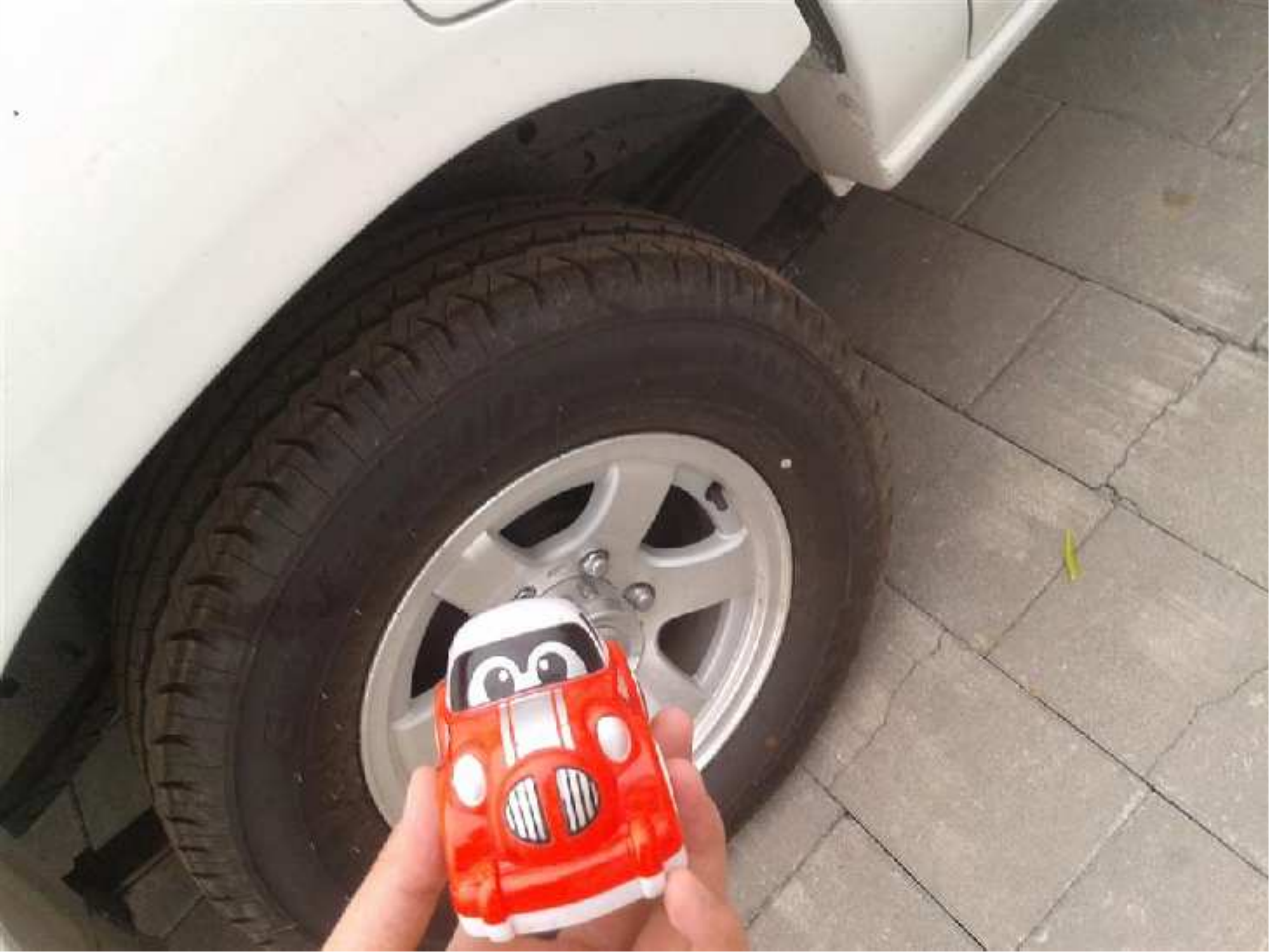} &
\includegraphics[height=1.8cm]{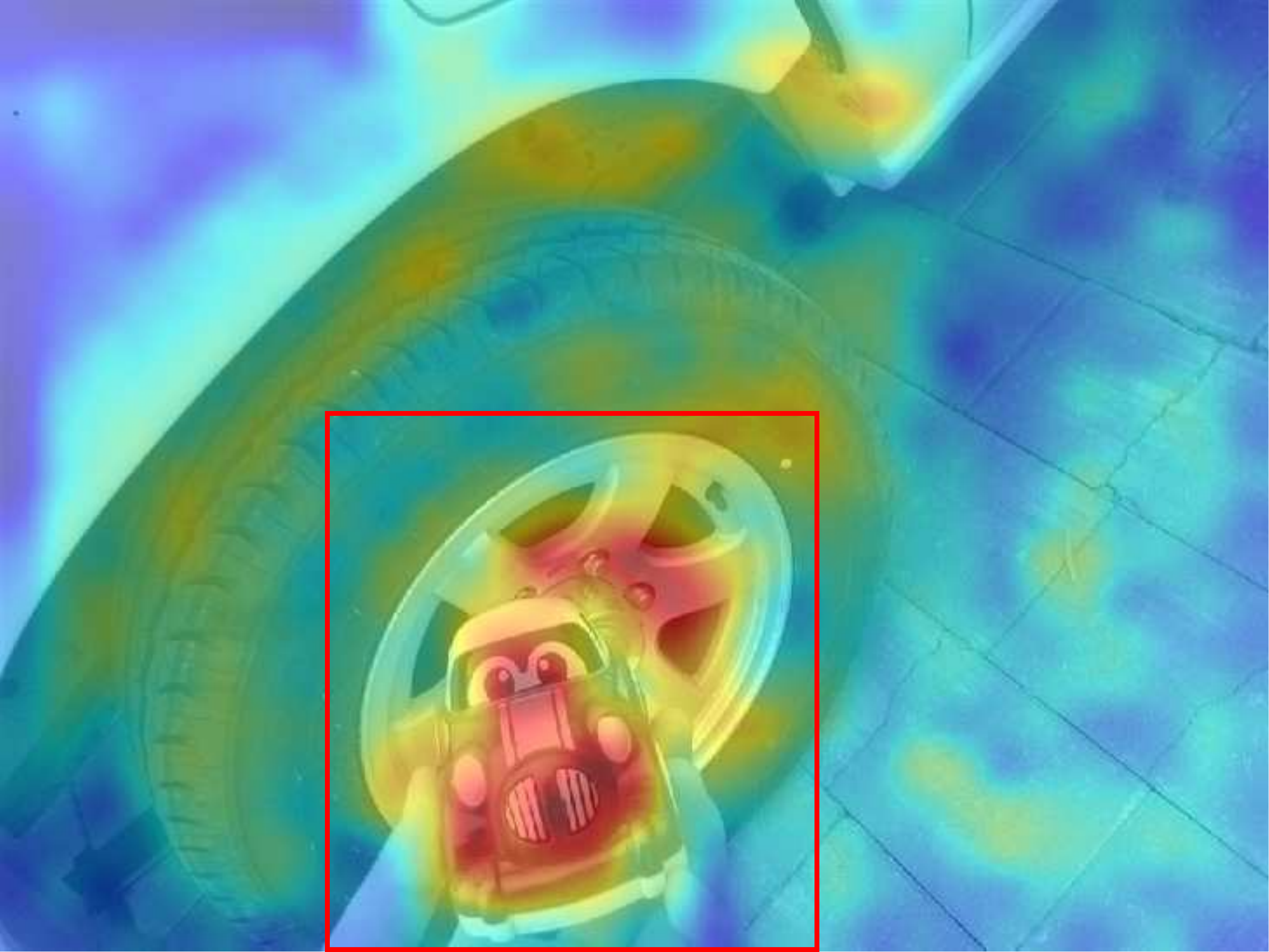} &
\includegraphics[height=1.8cm]{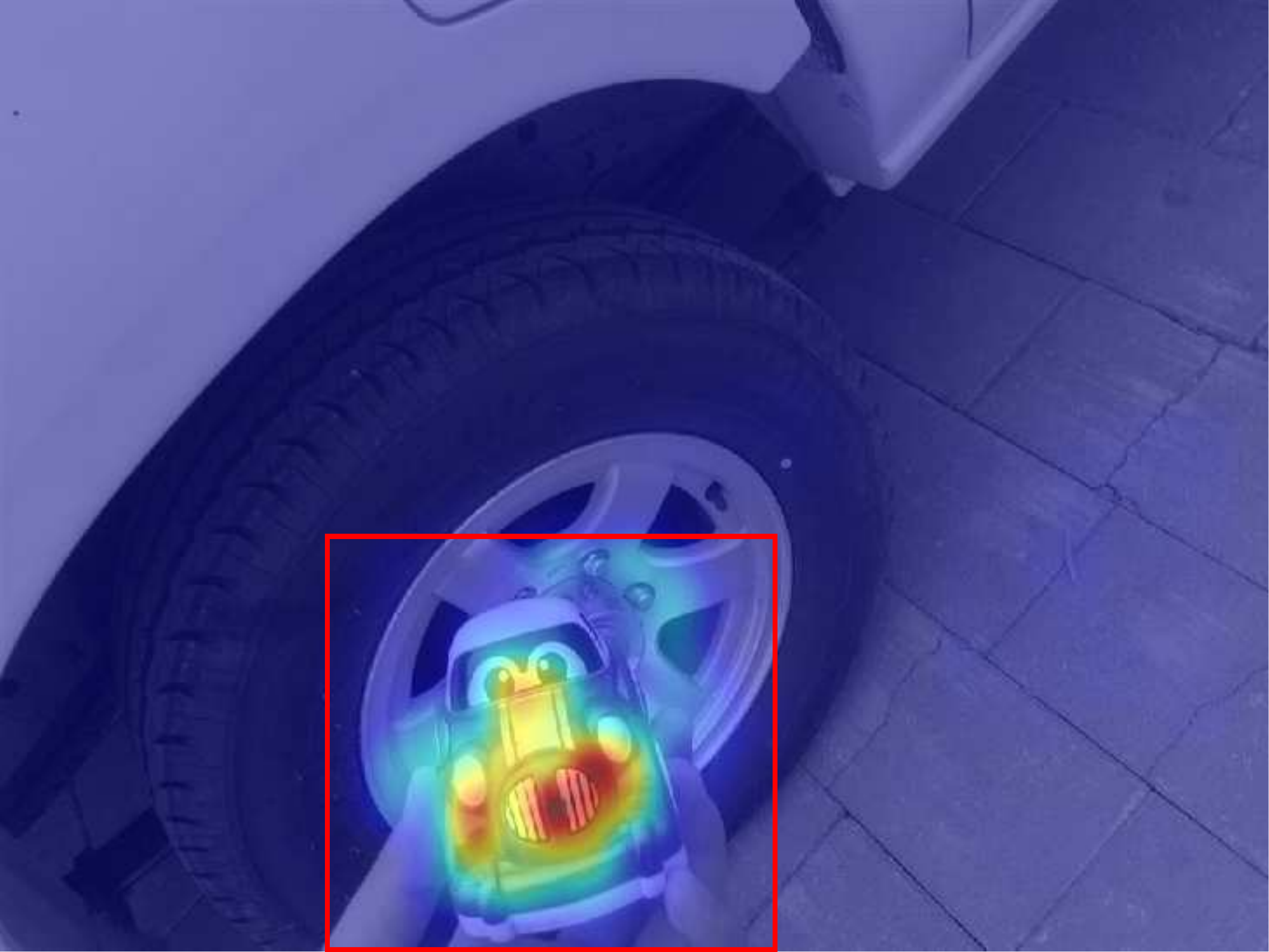} \\[2pt]
&&&&\\[-8pt]
{\normalsize image} &
{\normalsize \FSegm} &
{\normalsize \OSegm}
\\

  \end{tabular}
\end{center}

\caption{Examples of images from Oxford5k (first 2 rows) and Instre (last 3 rows) datasets, along with smoothed \FS and \OS maps superimposed on the images and regions detected by EGM, in red.
\vspace{-20pt}
}
\label{fig:example}
\end{figure}

We exploit the fact that Instre dataset comes with bounding box annotation for all database images.
We use the ground truth information to quantitatively evaluate the saliency maps.
We define \emph{precision} as the sum of saliency over ground truth regions, normalized by the sum over the entire image, and we measure it for \FS and \OS as shown in Figure~\ref{fig:regDet}.
High precision means that a saliency map is well aligned to the ground truth bounding boxes.
Given that these bounding boxes are not used anywhere, the improvement that \OS offers is impressive.
Visual examples for saliency maps and detections for \FSegm and \OSegm are shown in Figure~\ref{fig:example}. In all cases, OS is cleaner and focuses on objects that FS cannot discriminate.

\subsection{Comparison to other methods}

\begin{table}
\def \rmac{\scriptsize R\hspace{-0.7pt}-\hspace{-0.7pt}MAC}
\def \rcrow{\scriptsize R\hspace{-0.7pt}-\hspace{-0.7pt}CroW}

\def \edgeBox{\scriptsize Edge\hspace{-0.7pt}-\hspace{-0.7pt}Box}
\def \RPN{\scriptsize RPN}

\def \aqe{\scriptsize AQE~\cite{CPSIZ07}}
\def \scsm{\scriptsize SCSM~\cite{SLBW14}}
\def \hn{\scriptsize HN~\cite{DGBQG11}}
\def \hqe{\scriptsize HQE}
\def \crow{\scriptsize CroW~\cite{KMO15}}
\def \netvlad{\scriptsize NetVLAD~\cite{AGT+15}}
\def \rmatch{\scriptsize R\hspace{-0.7pt}-\hspace{-0.7pt}match~\cite{RSAC14}}
\def \reg{\scriptsize Diffusion~\cite{ITA+16}}
\def \glob{\scriptsize Diffusion~\cite{ITA+16}}

\footnotesize
\begin{center}
\setlength\extrarowheight{2pt}
\small
\resizebox{0.4\textwidth}{!}{\begin{tabular}{ |@{\msp}l@{\msp}|@{\msp}l@{\msp}|@{\msp}c@{\msp}|@{\msp}c@{\msp}|@{\msp}c@{\msp}|}
    \hline
      Method                                                    & QE          & Instre    & Oxford          & Oxford105k  \\ \hline \hline
      MAC                                                       & -           & 48.5      & 79.7            & 73.9        \\
      Uniform~\cite{TSJ15}                                      & -           & 47.7      & 77.7            & 70.1        \\
      \FSegm$^{\star}$                                          & -           & 48.4      & 77.5            & 70.2        \\
      \OSegm$^{\star}$                                          & -           & 50.1      & 79.6            & 71.8        \\
      \OSegm-${\triangle}^\star$                                & -           & 53.7      & 79.8            & 71.4        \\  \hline \hline
      MAC                                                       & \checkmark  & 71.8      & 87.4            & 86.0        \\
      Uniform~\cite{TSJ15}                                      & \checkmark  & 70.3      & 85.7            & 82.7        \\
      \FSegm$^{\star}$                                          & \checkmark  & 71.2      & 89.8            & 87.9        \\
      \OSegm$^{\star}$                                          & \checkmark  & 72.7      & \os{90.4}       & \os{88.0}   \\
      \OSegm-$\triangle^\star$                                  & \checkmark  & \os{75.4} & 90.1            & 84.3        \\  \hline
\end{tabular}}

\caption{mAP comparison of our methods marked with $^\star$ against baselines on all tested datasets. QE refers to query expansion by diffusion~\cite{ITA+16}.
\vspace{-20pt}
}
\label{tab:soa}
\end{center}
\end{table}

We compare our methods to the standard practice of uniform region sampling (Uniform) as in R-MAC and global max pooling (MAC).
We additionally propose a variant of \OSegm, where further uniform region sampling at 2 scales is performed within each detected region.
We refer to this as \OSegm-$\triangle$.
All methods are tested with $k$-NN search and global diffusion~\cite{ITA+16}, which is a method for query expansion or manifold search and is known to significantly improve performance.
Results are given in Table~\ref{tab:soa}.

\FSegm improves performance compared to uniform sampling by focusing on salient objects.
However, salient objects are not necessarily relevant for the particular dataset.
This is what \OSegm captures and boosts the search performance, especially on Instre.
On all datasets, MAC is better than uniform sampling (R-MAC). This is because the network used~\cite{RTC16} is directly fine-tuned to optimize MAC.
However, when using diffusion, we outperform it on all datasets.
This can be explained by the fact that diffusion boosts any items that are similar to the top-ranking ones according to the original similarity~\cite{ITA+16}, so it is essential that these items are reliable. A global descriptor is affected by clutter in general. By contrast, our representation is global yet clutter-free.
Our improvements are larger on Instre, which is more challenging due to small objects and severe background clutter.
This is exactly where our detection is essential. Most Instre images are also quite different than the building images which the network is fine-tuned on. This is probably why our representations outperform MAC even without diffusion on this dataset.

There are several other previous approaches that deal with region detection or saliency masks, which are not directly comparable, so they are not included in Table~\ref{tab:soa}.
Nevertheless, we outperform their reported results.
Salvador \etal~\cite{SGMS16} use the off-the-shelf VGG and fine-tune RPN in the test set.
Without using query expansion, they obtain $71.0$ in Oxford5k.
Similarly, Jimenez \etal~\cite{JAG17} learn class weights and apply them on the activation maps of off-the-shelf VGG and achieve $73.6$ in Oxford5k.
Song \etal~\cite{SHGXS17} train on different datasets, and achieve $78.3$ in Oxford5k.
The results obtained by learning a saliency mask are not comparable since spatial verification with local features is always applied in the end~\cite{NASH16}.
Finally, Zheng \etal~\cite{ZWW+16} achieve 83.4 with regional representation on Oxford5k.
They employ both CNN and local features, while we only rely on CNN and much more compact representation.
Finally, no work other than~\cite{ITA+16} evaluates on Instre which is rather challenging due to small objects.

\begin{figure}
\centering
\input{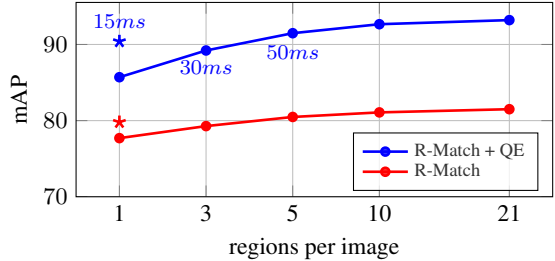}
\small
\begin{tabular}{c}
\extfig{uniReg}{
\begin{tikzpicture}
\begin{axis}[%
	width=0.95\linewidth,
	height=0.5\linewidth,
	xlabel={regions per image},
	ylabel={mAP},
	legend cell align={left},
	legend pos=south east,
    legend style={cells={anchor=west}, font =\scriptsize, fill opacity=0.8, row sep=-2.5pt},
    xmin = 0,
	ymin = 70,
   	xtick={1,3,5,7,10},
 	xticklabels={1,3,5,10,21},
    grid=both,
]

	\addplot[color=blue,     solid, mark=*,  mark size=1.5, line width=1.0] table[x=reg, y expr={100*\thisrow{regWhitenOxf5k}}] \diffUniformReg;\leg{R-Match + QE};
		\addplot[color=red,     solid, mark=*,  mark size=1.5, line width=1.0] table[x=reg, y expr={\thisrow{regWhitenOxf5k}}] \uniformReg;\leg{R-Match};
\addplot[color=blue, mark=star, only marks, mark size = 2.5, line width = 1] coordinates {(1,90.4)};
\addplot[color=red, mark=star, only marks, mark size = 2.5, line width = 1] coordinates {(1,79.8)};
	\node [above] at (axis cs:  1,  91) {\footnotesize \textcolor{blue}{$15 ms$}};
	\node [below] at (axis cs:  3,  89) {\footnotesize \textcolor{blue}{$30 ms$}};
	\node [below] at (axis cs:  5,  91) {\footnotesize \textcolor{blue}{$50 ms$}};

\end{axis}
\end{tikzpicture}
}

\end{tabular}
\caption{mAP comparison of our global \OSegm ($\star$) to R-Match with uniformly sampled regional descriptors, with and without diffusion on Oxford5k.
Text labels refer to query time.
\label{fig:uniReg}
\vspace{-10pt}
}
\end{figure}

\medskip
\head{Region cross-matching} methods~\cite{RSAC14} represent an image with multiple vectors, sacrificing memory footprint and complexity for accuracy.
In particular, the memory is linear in the number of regions, while the complexity is quadratic.
We compare our global representation with region cross-matching (R-Match) and regional diffusion~\cite{ITA+16} in Figure~\ref{fig:uniReg}.
Different numbers of regions are obtained by GMM reduction, exactly as in~\cite{ITA+16}.

Compared to regional descriptors, we require about $4$ times less memory to achieve the same performance.
The runtime complexity gain is in the order of $4^2$, which holds for the case of R-Match and also for the first part of diffusion where Euclidean nearest neighbors are found.
The diffusion complexity is O($m$), where $m$ is the number of non-zero entries of the graph.
We found that $m$ is 3.7 times smaller in our case and our measurements of actual query timings agree with this ratio.

\section{Conclusions}
\label{sec:discussion}
We propose a region detection approach that is dataset specific but requires no supervision.
It captures not only salient objects by considering each image individually but also frequently appearing ones by considering the dataset as a whole.
As a result, we avoid separate indexing of regional descriptors and construct a global descriptor by pooling over data-dependent regions, which performs well under background clutter and severe occlusions.
We demonstrate that this approach is effective in particular object retrieval where background clutter is a common problem.

\paragraph{Acknowledgments}
The authors were supported by the MSMT LL1303 ERC-CZ grant. The Tesla K40 used for this research was donated by the NVIDIA Corporation.

{\small
\bibliographystyle{ieee}
\bibliography{tex/egbib}
}

\end{document}